\documentclass[10pt,twocolumn,letterpaper]{article}

\usepackage{cvpr}              % To produce the CAMERA-READY version
\usepackage{soul}
\usepackage{bm}
\definecolor{cvprblue}{rgb}{0.21,0.49,0.74}
\usepackage[pagebackref,breaklinks,colorlinks,allcolors=cvprblue]{hyperref}

\def\confName{CVPR}
\def\confYear{2025}

\usepackage{multirow}
\usepackage{wrapfig}
\usepackage{floatrow}
\usepackage{xcolor}
\usepackage{listings}
\usepackage{color}
\usepackage{colortbl}
\usepackage{tcolorbox}
\usepackage{adjustbox}
\usepackage{floatrow}
\newfloatcommand{capbtabbox}{table}[][\FBwidth]
\title{\raisebox{-0.27\height} {\includegraphics[width=0.3in]{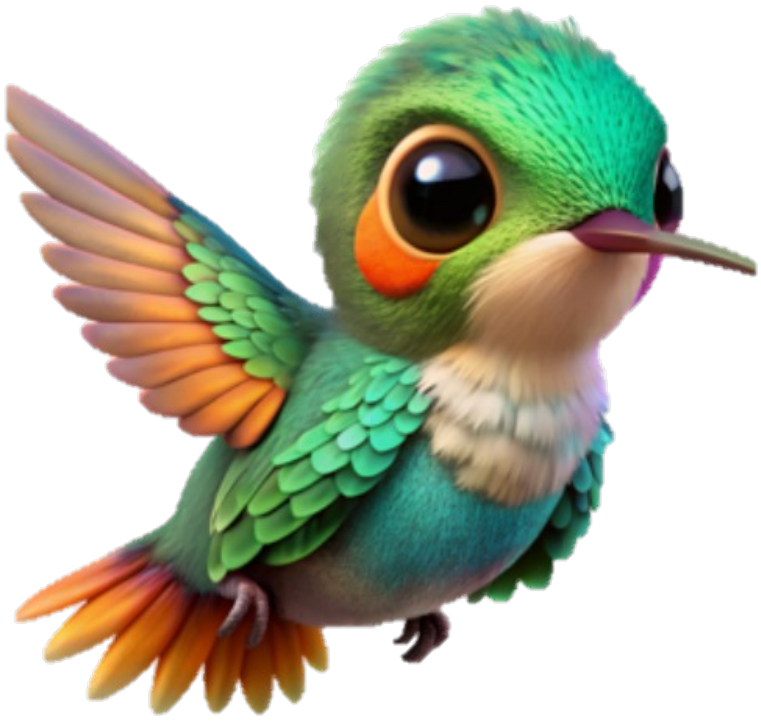}}\textit{LLaVA-ST}: A Multimodal Large Language Model for Fine-Grained Spatial-Temporal Understanding}

\author{
\textbf{Hongyu Li\textsuperscript{1}\footnotemark[1]}, \quad
\textbf{Jinyu Chen\textsuperscript{1}\footnotemark[1]}, \quad
\textbf{Ziyu Wei\textsuperscript{1}\footnotemark[1]},  \quad
\textbf{Shaofei Huang\textsuperscript{2,3}}, \quad
\textbf{Tianrui Hui\textsuperscript{2}}, \\
\textbf{Jialin Gao\textsuperscript{4}\footnotemark[2]~}, \quad
\textbf{Xiaoming Wei\textsuperscript{4}}, \quad
\textbf{Si Liu\textsuperscript{1}\footnotemark[2]~}\\
\textbf{$^{1}$}School of Artificial Intelligence, Beihang University \\
\textbf{$^{2}$}School of Computer Science and Information Engineering, Hefei University of Technology \\
\textbf{$^{3}$}Institute of Information Engineering, Chinese Academy of Sciences\\
\textbf{$^{4}$}Meituan \\
\tt\footnotesize \{hongyuer0317, nowherespyfly, huitianrui\}@gmail.com~~~\{liusi, chenjinyu, weiziyu\}@buaa.edu.cn\\
\tt\footnotesize \{gaojialin04, weixiaoming\}@meituan.com \\
}

\begin{document}

\twocolumn[{
\renewcommand\twocolumn[1][]{#1}
\maketitle
\begin{center}
    \centering
    % \vspace*{-.8cm}
    \includegraphics[width=1.01\textwidth]{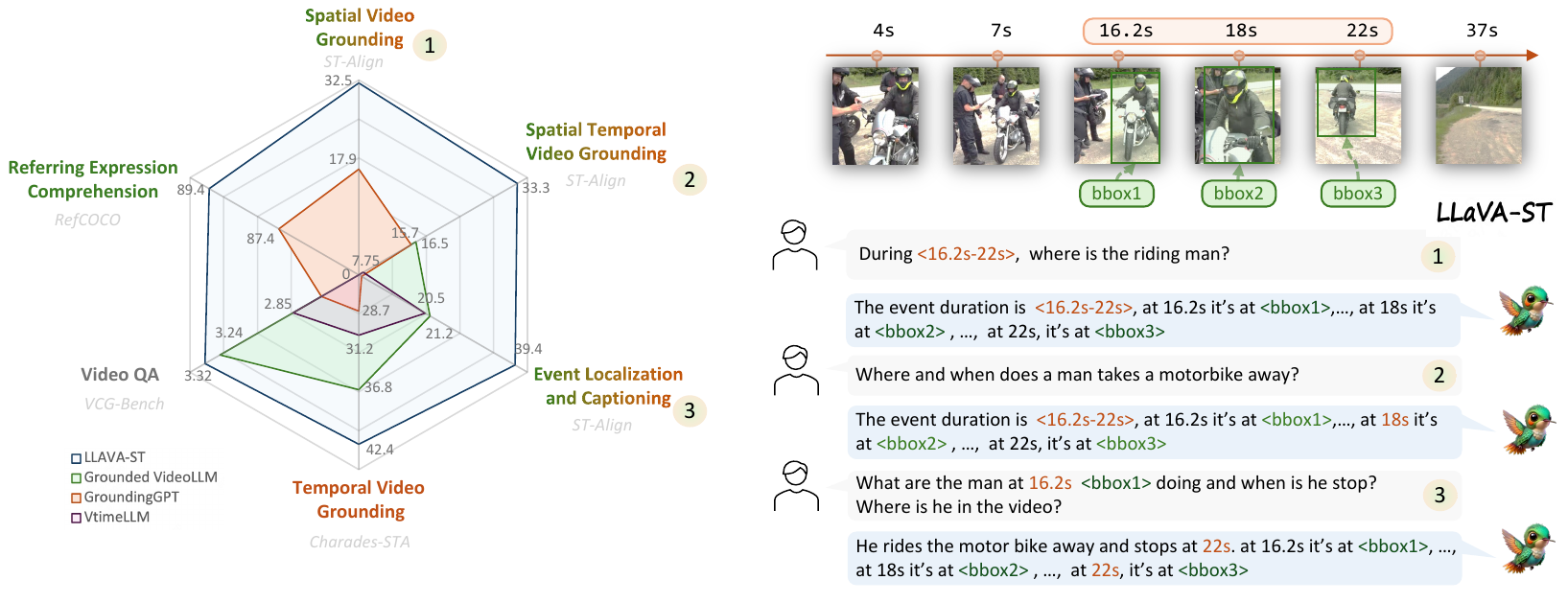}
    % \vspace*{-.2cm}
    \captionof{figure}{(Left) \textit{LLaVA-ST} demonstrates high performance across various tasks of fine-grained multimodal understanding and is the first MLLM capable of simultaneously processing spatial-temporal fine-grained understanding tasks. (Right) Examples of spatial-temporal interleaved fine-grained understanding tasks in the proposed ST-Align, which include Spatial Temporal Video Grounding (STVG), Event Localization and Captioning (ELC), and Spatial Video Grounding (SVG).}
\label{fig:fig1}
\end{center}
}]

\let\thefootnote\relax\footnotetext{
$^*$ Equal contribution \hspace{5pt}$^\dagger$ Corresponding author.
}

\label{sec:intro}

\begin{abstract}
Recent advancements in multimodal large language models (MLLMs) have shown promising results, yet existing approaches struggle to effectively handle both temporal and spatial localization simultaneously. 
This challenge stems from two key issues: first, incorporating spatial-temporal localization introduces a vast number of coordinate combinations, complicating the alignment of linguistic and visual coordinate representations; second, encoding fine-grained temporal and spatial information during video feature compression is inherently difficult.
To address these issues, we propose \textit{LLaVA-ST}, a MLLM for fine-grained spatial-temporal multimodal understanding.
In \textit{LLaVA-ST}, we propose Language-Aligned Positional Embedding, which embeds the textual coordinate special token into the visual space, simplifying the alignment of fine-grained spatial-temporal correspondences. 
Additionally, we design the Spatial-Temporal Packer, which decouples the feature compression of temporal and spatial resolutions into two distinct point-to-region attention processing streams.
Furthermore, we propose ST-Align dataset with 4.3M training samples for fine-grained spatial-temporal multimodal understanding.
With ST-align, we present a progressive training pipeline that aligns the visual and textual feature through sequential coarse-to-fine stages.Additionally, we introduce an ST-Align benchmark to evaluate spatial-temporal interleaved fine-grained understanding tasks, which include Spatial-Temporal Video Grounding (STVG) , Event Localization and Captioning (ELC) and Spatial Video Grounding (SVG). \textit{LLaVA-ST} achieves outstanding performance on 11 benchmarks requiring fine-grained temporal, spatial, or spatial-temporal interleaving multimodal understanding. Our code, data and benchmark will be released at \href{https://github.com/appletea233/LLaVA-ST}{https://github.com/appletea233/LLaVA-ST}.
\end{abstract}    
\section{Introduction}
Multimodal Large Language Models (MLLMs) have recently made significant progress in multimodal understanding~\cite{elysium,groundedvideollm,llavaov,blip2}, demonstrating strong visual comprehension abilities~\cite{vtimellm,visionllm,groundinggpt,llamaadapter}.  
Beyond generating pure language responses about visual content, some MLLMs are tasked with more complex challenges—specifically, understanding and outputting the coordinates of events or visual entities, \eg, Temporal Video Grounding (TVG)~\cite{vtimellm,groundedvideollm,groundinggpt} and Referring Expression Comprehension (REC)~\cite{chen2023shikra,ma2023vista,ma2025groma}. In this paper, \textbf{we refer to tasks that require processing visual coordinates based on linguistic input as fine-grained multimodal understanding.}
For these tasks, current MLLMs primarily focus on two aspects: models like~\cite{chatuniv,chen2023shikra} excel at spatial localization of objects within images but struggle with fine-grained temporal understanding tasks; models like~\cite{videochat,videollava,lita,vtimellm} specialize in fine-grained temporal understanding like TVG, yet are incapable of determining bounding boxes for objects. Existing MLLMs cannot uniformly achieve spatial, temporal and interleaved fine-grained multimodal understanding tasks.

Designing a unified MLLM for spatial-temporal fine-grained understanding tasks faces two major challenges:
i) \textbf{Difficulties in multimodal coordinate alignment}. When MLLMs perform joint spatial-temporal fine-grained understanding, the coordinate space expands, making cross-modal coordinate alignment more complex. This challenge is compounded by the lack of datasets that provide spatiotemporal tubes paired with corresponding linguistic descriptions; most existing datasets typically provide only bounding boxes or temporal durations. As a result, effectively localizing objects in both spatial and temporal coordinates becomes increasingly difficult.
ii) \textbf{Challenges in preserving visual details}.
To alleviate the computational burden, MLLMs are required to compress the substantial feature extracted from video inputs. However, this compression must not only preserve spatial and temporal relationships but also maintain fine-grained contextual information, all within a limited token budget.
Naive compression methods, such as Q-Former~\cite{blip2,groundinggpt} or pooling~\cite{groundedvideollm,xu2024slowfast}, lead to an inevitable loss of spatial relationships and fine-grained details, which hinders the simultaneous comprehension from both spatial and temporal aspects.

\begin{figure*}[ht]
    \centering
    \includegraphics[width=0.9\linewidth]{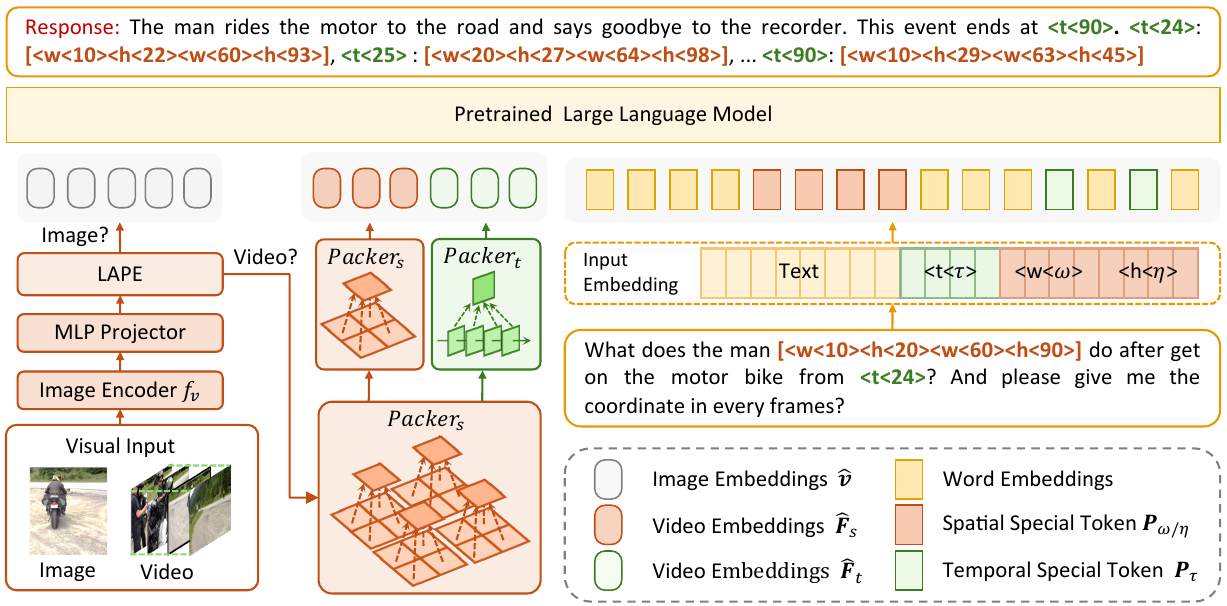}
    \caption{The Overall Architecture of \textit{LLaVA-ST}. In \textit{LLaVA-ST}, we introduce discrete special tokens to represent spatio-temporal coordinates within the language modality. LAPE embed these coordinate representations into the visual feature space. Furthermore, the STP module utilizes a two-stream packing mechanism to efficiently compress the features.}
    \label{fig:method}
\end{figure*}

To address the above challenges, we propose \textit{LLaVA-ST}, an MLLM for fine-grained spatial-temporal understanding. 
\textit{LLaVA-ST} is the first MLLM capable of simultaneously understanding and localizing the temporal range and spatial coordinates of visual inputs. 
To alleviate the difficulty of cross-modal coordinate alignment, we propose Language Aligned Positional Embedding (LAPE), which directly embeds the features representing the spatial and temporal coordinates in the textual space as the positional embeddings for visual features, thus bypassing the difficult cross-modal coordinate alignment. 
To achieve efficient video compression, we propose the Spatial-Temporal Packer (STP), which separates temporal and spatial compression processes, maintaining more spatial-temporal relationships. 
Additionally, it employs a region-to-point attention mechanism to compress visual features, preserving more fine-grained information.

To enable \textit{LLaVA-ST} to support various fine-grained multimodal understanding tasks, we apply a progressive training strategy consisting of three sequential stages: content alignment, coordinate alignment, and multi-task instruction tuning. In addition, we collect and curate a dataset of $4.3$M samples, including 15 types of fine-grained multimodal understanding data, named ST-Align. Within ST-Align, we collect $228$K new instruction tuning samples for spatial temporal interleaved fine-grained understanding tasks, which include Spatial-Temporal Video Grounding (STVG)~\cite{Vidstg} and two newly introduced tasks: Event Localization and Captioning (ELC) and Spatial Video Grounding (SVG). 
Furthermore, we construct corresponding benchmarks to evaluate the capabilities of MLLMs. 
Based on the aforementioned training strategies and data, \textit{LLaVA-ST} exhibits strong multimodal understanding capabilities and achieves state-of-the-art performance on multiple benchmarks, as shown in Fiugre~\ref{fig:fig1}.

Our contributions are summarized as follows:
i) We propose \textit{LLaVA-ST}, the first MLLM capable of end-to-end processing fine-grained spatial, temporal, and interleaved fine-grained multimodal understanding tasks. In \textit{LLaVA-ST}, we introduce the LAPE, which reduces the difficulty of aligning coordinate features between vision and language. Additionally, we propose STP, which preserves fine-grained spatiotemporal context in the video feature compression process.
ii) We propose a progressive training strategy for \textit{LLaVA-ST}, enabling the model to progressively learn content alignment, coordinate alignment and multi-task ability. To support the training process, we construct the ST-Align dataset which includes 15 different tasks and 228K newly conducted data samples.
iii) We perform experiments for on 11 different benchmarks, including TVG, video QA, REC and spatial-temporal interleaved tasks,  and the results demonstrate the outstanding capabilities of \textit{LLaVA-ST}.

\section{Related Works}
\label{sec:related_works}
\subsection{Multi-Modal Understanding tasks}
Current tasks  for MLLMs can be broadly classified into four categories.
The first category is \emph{general visual understanding}, where models generate dialogues or descriptions based on images or videos, such as Video Captioning (VC)~\cite{webvid,panda70m,YouCook2,sharegpt4video}  and Visual Question Answering (VQA)~\cite{ego4d,clever,webvid-qa}.
The second category requires \textit{fine-grained spatial understanding} of images and videos, necessitating the ability to handle specific spatial coordinates of visual content. Examples include Referring Expression Comprehension (REC)~\cite{refcoco,refcocog,huang2020referring,hui2020linguistic},, Dense Grounded Captioning (DGC)~\cite{glamm}, Panoptic Narrative Grounding (PNG)~\cite{gonzalez2021panoptic,li2024dynamic,hui2023enriching,ding2022ppmn} and Referring Video Object Segmentation (RVOS) ~\cite{seo2020urvos,huang2024unleashing,hui2021collaborative,hui2023language}, .
The third category demands \textit{fine-grained temporal understanding} of videos, enabling localization and comprehension of temporal ranges within videos. This includes tasks like TVG~\cite{vtimellm,didemo,vtgllm}, Dense Video Captioning (DVC)~\cite{coin,vitt,YouCook2}, and Temporal Grounded Conversation (TGC)~\cite{ANet-cap,momentor,groundedvideollm}.
The fourth category requires \textit{simultaneous fine-grained spatial and temporal understanding} of videos, allowing models to localize and process spatiotemporal tubes in videos, such as in Spatio-Temporal Video Grounding~\cite{Vidstg}. Existing MLLMs are primarily capable of handling only one or two different categories of tasks. We expand the tasks that require simultaneous fine-grained spatiotemporal understanding for MLLMs in ST-Align, including STVG, ELC, and SVG.

\subsection{Multimodal Large Language Models}
Current MLLMs have demonstrated substantial advancements in a variety of vision and language tasks. Methods such as those proposed in \cite{videollama,video-chatgpt,ma2023vista,moviechat,kong2025controllable} are capable of generating visual captions or answering questions pertaining to visual content. Certain models are specifically designed to achieve spatial fine-grained understanding. \cite{you2023ferret,wang2022ofa,chen2023shikra,elysium} is adept at understanding referring expressions and provides fine-grained spatial grounding results inputs. On another front, MLLMs in \cite{vtimellm,lita,momentor} are effective in performing temporal grounding tasks in lengthy videos. However, existing MLLMs face challenges in achieving simultaneous fine-grained spatial-temporal understanding. Moreover, they struggle with tasks that require spatial-temporal interleaved localization. GroundingGPT~\cite{groundinggpt} supports temporal localization in videos and referential grounding in images. However, it is unable to process spatial-temporal interleaved tasks in an end-to-end manner. Unlike existing models, \textit{LLaVA-ST} unifies the end-to-end processing of the spatial-temporal fine-grained understanding tasks and demonstrates superior abilities.

\section{Model Architecture}

\subsection{Overview}

The overall architecture of our \textit{LLaVA-ST} is illustrated in Figure~\ref{fig:method}, where both video and image can be processed.
For the video $V$ consisting of $T$ frames, we sample $N$ frames $\bm{I}_{1:N}$ at equal intervals. $N$ is set to $1$ for the image input.

A visual encoder $f_v$ then sequentially extracts features from each sampled frames. These features are mapped into the textual space using a Multi-Layer Perceptron (MLP):
\begin{equation}
    \bm{v} = \text{MLP}(f_v(\bm{I}_{1:N})).
\end{equation}
where $\bm{v} \in \mathbb{R}^{W_1 \times H_1 \times D}$, and $W_1$ and $H_1$ represent the spatial resolution $f_v$'s output. Subsequently, we employ Language-Aligned Positional Embedding (LAPE) to overlay language coordinate feature $\hat{\bm{\rho}}$ into $\bm{v}$:
\begin{equation}
    \hat{\bm{v}} = \bm{v} + \hat{\bm{\rho}}.
\end{equation}
For image inputs, the resulting $ \hat{\bm{v}} $ can be directly flatten and fed into the LLM for processing. In the case of videos, we utilize Spatial-Temporal Packer (STP) to compress the features along temporal and spatial dimensions separately:
\begin{equation}
        \hat{\bm{F}}_s,\hat{\bm{F}_t} = \text{STP}(\bm{\hat{v}}), 
\end{equation}
The $ \hat{\bm{F}}_s $ and $ \hat{\bm{F}}_t $ are then flattened and input into the LLM~\cite{llama}. 
The LLM will leverage visual features and language embeddings to address fine-grained multimodal understanding tasks in a unified manner. We elaborate our LAPE and STP modules in \S\ref{sec:LAPE} and \S\ref{sec:STP}. 

\begin{figure}[t]
    \centering
    \includegraphics[width=\linewidth]{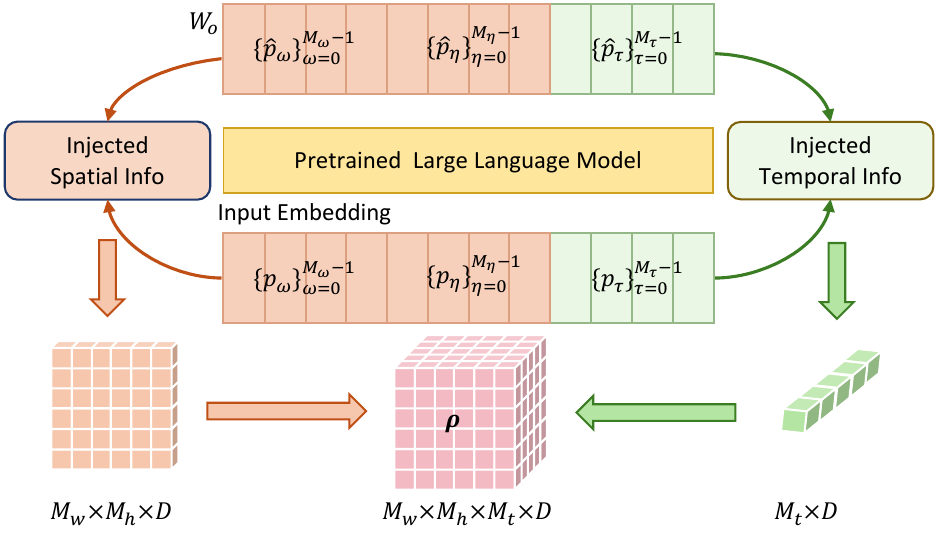}
    \caption{Details of the LAPE. LAPE leverages coordinate-related input text embeddings and features within the output layer matrix as visual positional embeddings.}
    \label{fig:LAPE}
\end{figure}
\subsection{Language-Aligned Positional Embedding}
\label{sec:LAPE}

To avoid the inefficiency of tokenizing numerical text, we extend the vocabulary of the LLM by introducing special tokens for coordinates. These special tokens are incorporated into the input text embeddings and output layers of the LLM. We then embed the spatial tokens into the visual features as positional embeddings, as shown in Figure~\ref{fig:LAPE}.

 \noindent\textbf{Definition of Special Tokens}. Specifically, we evenly divide the temporal dimension into $ M_t - 1 $ segments, obtaining $ M_t $ anchor points. The timestamp $t$ is represented by the $\tau$-th anchor point that is closest to its temporal position and encode to \texttt{<t<$\tau$>}.  This results in $ M_t $ special tokens along the temporal axis.
Similarly, we partition the video's width $ W $ and height $ H $ into $M_w - 1$ and $M_h - 1$ equal parts respectively, thereby obtaining $ M_w + M_h $ special tokens along the spatial dimensions. For the spatial coordinate $(w, h)$, their corresponding special tokens are \texttt{<w<$\omega$>}, \texttt{<h<$\eta$>}.

\noindent\textbf{Input/Output of Special Tokens}.
To directly incorporate these special tokens into the LLM, we extend the original input text embeddings within the LLM so that each special token is encoded as an individual word embedding. The text input embeddings of \texttt{<t<$\tau$>}, \texttt{<w<$\omega$>} and \texttt{<h<$\eta$>} are denoted as $\bm{p}_\omega$, $\bm{p}_\eta$ and $\bm{p}_\tau$.
 Similarly, we also expand the parameter matrix in the final-layer of the LLM to support the coordinate special tokens. The output layer of LLM is:
\begin{equation}
    \bm{o}_l = \text{softmax}( \bm{W}_o \times \bm{h}_l^{\top}),
\end{equation}
where $\bm{W}_o \in \mathbb{R}^{(V+M_w+M_h+M_t)\times D}$ is the parameter, $V$ is the vocabulary size for language, $\bm{h}_l\in\mathbb{R}^{1\times D}$ is the text feature, and $\bm{o}_l$ is the predicted next word possibility. Consequently, in this process, the model essentially computes the cosine similarity between  $\bm{h}_l$ and the row vectors of $\bm{W}_o$. When the cosine similarity is high, the probability of the model outputting that particular word increases correspondingly. Therefore, row vectors within $\bm{W}_o$ can also serve as effective representations of the respective words. Let $\hat{\bm{p}}_\omega$, $\hat{\bm{p}}_\eta$, and $\hat{\bm{p}}_\tau$ denote the row vectors in $\bm{W}_o$ corresponding to \texttt{<w<$\omega$>}, \texttt{<h<$\eta$>}, and \texttt{<t<$\tau$>}, respectively.

\setlength{\fboxsep}{0pt}
\begin{table*}
\resizebox{\linewidth}{!}{
    \begin{tabular}{l|c|c|c}
    \toprule
    \textbf{Training Stage} & \textbf{Task} &\textbf{\# of Samples} & \textbf{Data Source}\\
    \midrule 
    Content Alignment (Stage-1) & \multirow{1}{*}{Video Captioning} &1.28M &WebVid-10M~\cite{webvid}, Panda-70M~\cite{panda70m}, InternVid-10M~\cite{internvideo2}  \\
    \midrule 
    \multirow{6}{*}{Coordinate Alignment (Stage-2) } 
    & \cellcolor{orange!10} Temporal Video Grounding & \cellcolor{orange!10} 149K & \cellcolor{orange!10} VTimeLLM-Stage2~\cite{vtimellm}  \\
    & \cellcolor{orange!10} Dense Video Captioning & \cellcolor{orange!10} 92K & \cellcolor{orange!10} VTimeLLM-Stage2~\cite{vtimellm}, Moment-10M~\cite{momentor}, InternVid-G~\cite{internvideo2} \\
    & \cellcolor{orange!10} Temporal Referring & \cellcolor{orange!10} 95K & \cellcolor{orange!10} VTimeLLM-Stage2~\cite{vtimellm}, InternVid-G~\cite{intervid10m} \\
    & \cellcolor{blue!10} Referring Expression Comprehension & \cellcolor{blue!10} 101K & \cellcolor{blue!10} GranD~\cite{glamm}* \\
    & \cellcolor{blue!10} Dense Grounded Captioning & \cellcolor{blue!10} 150K & \cellcolor{blue!10} GranD~\cite{glamm}* \\
    & \cellcolor{blue!10} Region Caption & \cellcolor{blue!10} 100K & \cellcolor{blue!10} GranD~\cite{glamm}* \\
    \midrule 
    \multirow{14}{*}{Multi-Task Instruction Tuning (Stage-3) } 
    & \cellcolor{orange!10} Temporal Grounded Conversation & \cellcolor{orange!10} 459K & \cellcolor{orange!10} ANet-RTL~\cite{lita}, Moment-10M~\cite{momentor},  Grounded-VideoLLM~\cite{groundedvideollm}  \\
    & \cellcolor{orange!10} Temporal Video Grounding  & \cellcolor{orange!10} 84K &\cellcolor{orange!10} DiDeMo~\cite{didemo}, HiREST~\cite{HiREST}, QuerYD~\cite{queryd}, VTG-IT~\cite{vtgllm} \\
    & \cellcolor{orange!10} Dense Video Caption & \cellcolor{orange!10} 41K & \cellcolor{orange!10} COIN~\cite{coin}, ViTT~\cite{vitt}, YouCook2~\cite{YouCook2}, VTG-IT~\cite{vtgllm} \\
    & \cellcolor{blue!10} Image Grounded Conversation & \cellcolor{blue!10} 190K & \cellcolor{blue!10} MUSE~\cite{pixellm}*, Flickr30k Entities ~\cite{flickr30k}\\
    & \cellcolor{blue!10} Referring Expression Comprehension & \cellcolor{blue!10} 288K & \cellcolor{blue!10} RefCOCO~\cite{refcoco}, RefCOCO+~\cite{refcoco}, RefCOCOg~\cite{refcocog} \\
    & \cellcolor{blue!10} Dense Grounded Captioning & \cellcolor{blue!10} 22K & \cellcolor{blue!10} Flickr30k Entities ~\cite{flickr30k} \\
    & \cellcolor{blue!10} Region Caption & \cellcolor{blue!10} 288K & \cellcolor{blue!10} RefCOCO~\cite{refcoco}, RefCOCO+~\cite{refcoco}, RefCOCOg~\cite{refcocog} \\
    & \cellcolor{purple!10} Spatial-Temporal Video Grounding & \cellcolor{purple!10} 81K & \cellcolor{purple!10} Self collected \\
    & \cellcolor{purple!10} Event Localization and Captioning & \cellcolor{purple!10} 36K & \cellcolor{purple!10} Self collected \\
    & \cellcolor{purple!10} Spatial Video Grounding & \cellcolor{purple!10} 81K & \cellcolor{purple!10} Self collected \\
    & Converstation &222K &VCG-Plus-112K~\cite{videogpt+}, Videochatgpt-100K~\cite{VideoChatGPT}, Videochat2-Conv~\cite{2023videochat}\\
    & VideoQA &338K &EgoQA~\cite{ego4d}, NExT-QA~\cite{nextqa}, Intent-QA~\cite{intentqa}, AGQA~\cite{agqa}, STAR~\cite{star}, CLEVRER~\cite{clever}, WebVid-QA~\cite{webvid-qa}\\
    & Classification &66K &SthSthV2~\cite{sthsth}, Kinetics~\cite{kinetics}\\   
    & Video Captioning &168K &TextVR~\cite{textvr}, YouCook2~\cite{YouCook2}, WebVid~\cite{webvid}, ShareGPT4Video~\cite{sharegpt4video}\\   
                                          
    \bottomrule 
\end{tabular}
}
\caption{ Overview of ST-Align dataset. Tasks highlighted in \colorbox{orange!10}{\strut orange} involve datasets on temporal fine-grained understanding; those in \colorbox{blue!10}{\strut blue} pertain to spatial fine-grained understanding; and those in \colorbox{purple!10}{\strut pink} correspond to spatiotemporal interleaved fine-grained understanding.}
\label{tab:data}
\end{table*}

\noindent\textbf{Visual Positional Embeddings}. Previous methods~\cite{visionllm,groundedvideollm} primarily train special tokens implicitly. However, when it comes to fine-grained spatial and temporal understanding, it is challenging to train specific tokens that map to the corresponding spatial or temporal coordinates due to the enlarged coordinate combination.
By utilizing the special tokens as positional embeddings for 
the visual feature, the difficulty to align textual coordinate representations with visual coordinates is significantly reduced. $\hat{\bm{p}}_{\omega/\eta/\tau}$ and $\bm{p}_{\omega/\eta/\tau}$ are randomly initialized in finetuning, so both need to align with the coordinate of the visual features. We use $\hat{\bm{p}}_{\omega/\eta/\tau}$ and $\bm{p}_{\omega/\eta/\tau}$ simultaneously to build position embeddings:
\begin{equation}
    \bm{\rho}\{\omega,\eta,\tau\} = \frac{ \bm{\hat{p}}_{\omega} + \bm{p}_{\omega}}{2} + \frac{ \bm{\hat{p}}_{\eta}+\bm{p}_{\eta}}{2} + \frac{\bm{\hat{p}}_{\tau}  +\bm{p}_{\tau}}{2},
\end{equation}
where $\bm{\rho}\in \mathbb{R}^{M_w \times M_h \times M_t \times D}$ and $\bm{\rho}\{\omega,\eta,\tau\}$ for the vector in position $(\omega,\eta,\tau)$ of $\bm{\rho}$. Since the shape of $\bm{\rho}$ differ from those of $\bm{v}$, we employ linear interpolation to downsample $\bm{\rho}$ to match the shape of $\bm{v}$, resulting in $\hat{\bm{\rho}}$. 
Since temporal relationships do not need to be considered for image inputs, we only downsampling $\bm{\rho}\{0:M_w-1,~0:M_h-1,~0\}$ to build $\hat{\bm{\rho}}$. 
$\hat{\bm{\rho}}$ serves as the positional embedding of $\bm{v}$. Moreover, the downsampling process facilitates the information transfer among the special token embeddings within the LLM, which facilitated the convergence.
\subsection{Spatial-Temporal Packer}
\label{sec:STP}
In STP, we compress video features through a two-step process. In the first step, we reduce the spatial resolution of $\hat{\bm{v}}$ to obtain $ \hat{\bm{F}}$, thereby decreasing computational burden for the following process. In the second step, we separately compress along the spatial and temporal dimensions of $ \hat{\bm{F}}$ to obtain $ \hat{\bm{F}}_s$ and $ \hat{\bm{F}}_t$, respectively.

In the first step, we partition $\bm{\hat{v}}$ along the spatial dimensions into $ k_1 \times k_1 \times N$ patches and resize them into a sequence of regional features $\bm{F}\in \mathbb{R}^{(k_1 \times k_1 \times N) \times (\frac{W_1}{k_1} \times \frac{H_1}{k_1}) \times D} $. We then perform average pooling within each patch of $\bm{F}$ to obtain the sequence of point features $\bm{F}'\in \mathbb{R}^{(k_1 \times k_1 \times N) \times 1 \times D}$. 
To imbue $\bm{F}'$ with detailed information from $\bm{F}$ , we take $ \bm{F}' $ as queries and the corresponding $ \frac{W_1}{k_1} \times \frac{H_1}{k_1} $ regional features in $\bm{F}$ as keys and values to perform attention, thereby obtaining the resolution-reduced features $\hat{\bm{F}}\in\mathbb{R}^{k_1 \times k_1 \times N \times D}$. We denote the above process as:
\begin{equation}
\hat{\bm{F}} = \text{packer}_s(\hat{\bm{v}}, k_1 \times k_1).  
\end{equation}
Compared to directly reducing spatial resolution through pooling, the point-to-region attention mechanism better preserves fine-grained information, thereby enhancing the model's ability to localize spatiotemporal coordinates.
In the second step, we employ a compression method similar to that of the first stage to separately compress the features along the temporal and spatial dimensions. For the compression along spatial dimensions, we apply:
\begin{equation}
\hat{\bm{F}}_s = \text{packer}_s(\hat{\bm{F}}, k_2 \times k_2),    
\end{equation}
where $\hat{\bm{F}}_s\in \mathbb{R}^{k_2\times k_2 \times N \times D}$. For the temporal dimension, we partition $\hat{\bm{F}}$ into $k_1\times k_1 \times \sigma$ patches along the temporal dimension and employ the same pooling and attention strategies as in $\text{packer}_s(\cdot,\cdot)$. This process yields the temporal-dimension-reduced feature $ \hat{\bm{F}}_t$:
\begin{equation}
    \hat{\bm{F}}_t = \text{packer}_t(\hat{\bm{F}},\sigma), 
\end{equation}
where $\hat{\bm{F}}_t\in \mathbb{R}^{k_1\times k_1 \times \sigma \times D}$. And LLM will take the flattened $\hat{\bm{F}}_t$ and $\hat{\bm{F}}_s$ as input.

\section{ST-Align and Training Strategy}
\label{sec:training}
In this section, we introduce ST-Align, an MLLM training dataset designed for spatio-temporal fine-grained multimodal understanding, and detail the specific progressive training strategies for \textit{LLaVA-ST} based on ST-Align.
\subsection{ST-Align}
To train \textit{LLaVA-ST}, we propose ST-Align, which comprises approximately 4.3 million training samples, as shown in Table~\ref{tab:data}. To address the current deficiency of data involving fine-grained multimodal understanding with spatiotemporal interleaving, we employ \texttt{GPT-4-turbo} to revise and enhance the textual annotations of VidSTG~\cite{Vidstg}, adapting them for the following three tasks:
i) \textbf{Spatio-Temporal Video Grounding (STVG)}: localizing the spatiotemporal tube of the event subject in the video based on the event description.
ii) \textbf{Event Localization and Captioning (ELC)}: given a starting position and the bounding box of the event subject, locate the ending position of the event and provide a description of the event.
iii) \textbf{Spatial Video Grounding (SVG)}: based on the event duration and textual description, locate the tracklet of the event subject.
Additionally, we present 2,000 validation samples for each task to evaluate the fine-grained spatiotemporal understanding capabilities of the MLLM.

\begin{table*}[h]
  \centering
    \resizebox{\linewidth}{!}{
    \setlength{\tabcolsep}{1mm}{
\begin{tabular}{l|c|cccc|ccccc|ccc}
    \toprule
    \multirow{2}{*}{Model} &\multicolumn{1}{|c}{LLM} &\multicolumn{4}{|c}{Spatial-Temporal Video Grounding} &\multicolumn{5}{|c}{Event Localization and Captioning} &\multicolumn{3}{|c}{Spatial Video Grounding}\\
    \cmidrule(lr){3-6} \cmidrule(lr){7-11} \cmidrule(lr){12-14}
    &Scale &tIoU@0.5 &m$_t$IoU  &sIoU@0.5  &m$_s$IoU &tIoU$@0.5$ &m$_t$IoU &sIoU$@0.5$ &m$_s$IoU &METEOR & sIoU$@0.3$ &sIoU$@0.5$ &m$_s$IoU  \\
    \midrule 
    GroundingGPT  \citep{groundinggpt}  & 7B    & 7.1  & 12.2 & 2.9 & 9.2 & 4.8 & 6.6 & 2.1 & 6.4 & 8.2   & 19.7 & 5.4 & 17.9 \\
    VTimeLLM  \citep{vtimellm}  & 7B    & 7.1 & 15.5  & - & - & 33.1 & 40.3 & - & - & 6.0 &  - & - & -  \\
    Grounded-VideoLLM  \citep{groundedvideollm}  & 4B  & 30.0 & 33.0  & - & - & 53.1 & 56.4  & - & - & 7.2  & - & - & - \\
    \midrule 
    \textit{LLaVA-ST}   &7B    & \textbf{44.6} & \textbf{43.8} & \textbf{21.1} & \textbf{22.8} & \textbf{60.4} & \textbf{60.0} & \textbf{32.4} & \textbf{33.5} & \textbf{24.7} & \textbf{47.2} & \textbf{30.9} & \textbf{32.5}  \\  
    \bottomrule 
\end{tabular}
}}
\caption{Results on spatial-temporal interleaved fine-grained understanding tasks in ST-Align benchmark.}
\label{tab:results_stvg}
\vspace{-2mm}
\end{table*}

\begin{table*}
    
\begin{floatrow}
\capbtabbox{
\resizebox{0.95\linewidth}{!}{
\setlength{\tabcolsep}{7 pt}
 
\begin{tabular}{l|c|ccc|c}
    \toprule
    \multirow{2}{*}{Model} &\multicolumn{1}{|c}{LLM} &\multicolumn{4}{|c}{Charades-STA} \\
    \cmidrule(lr){3-6} 
    &Scale &R@0.3 &R@0.5 &R@0.7 &mIoU\\
    \midrule 
    Video-LLaMA \citep{videollama}  &7B   & 25.2 & 10.6 & 3.4  & \cellcolor{gray!10}{16.8} \\
    SeViLA \citep{sevila}            &3B   & 27.0 & 15.0 & 5.8  & \cellcolor{gray!10}{18.3} \\
    Video-ChatGPT \citep{video-chatgpt} &7B  & 27.2 & 6.2  & 1.9  & \cellcolor{gray!10}{19.7} \\
    Valley \citep{valley}            &7B    & 28.4 & 1.8  & 0.3  & \cellcolor{gray!10}{21.4} \\
    VideoChat2 \citep{mvbench}       &7B   & 38.0 & 14.3 & 3.8  & \cellcolor{gray!10}{24.6} \\
    VideoChat \citep{videochat}      &7B   & 32.8 & 8.6  & 0.0  & \cellcolor{gray!10}{25.9} \\
    Momenter \citep{momentor}        &7B  & 42.6 & 26.6 & 11.6 & \cellcolor{gray!10}{28.5} \\
    VTimeLLM \citep{vtimellm}        &7B  & 51.0 & 27.5 & 11.4 & \cellcolor{gray!10}{31.2} \\
    GroundingGPT~\cite{groundinggpt}  &7B  & - & 29.6 & 11.9 & \cellcolor{gray!10}{28.7}   \\
    TimeChat \citep{timechat}        &7B   & - & 32.2 & 13.4 & \cellcolor{gray!10}{-}   \\
    VTG-LLM \citep{vtgllm}           &7B   & -    & 33.8 & 15.7 & \cellcolor{gray!10}{-}    \\
    HawkEye \citep{hawkeye}          &7B  & 50.6 & 31.4 & 14.5 & \cellcolor{gray!10}{33.7}  \\
    Grounded-VideoLLM  \citep{groundedvideollm}  &4B    &  {54.2} &  {36.4} &  {19.7} & \cellcolor{gray!10}{{36.8}} \\  
    \midrule 
    \textit{LLaVA-ST}   &7B    & \textbf{63.1} & \textbf{44.8} & \textbf{23.4} & \cellcolor{gray!10}{\textbf{42.4}} \\  
    \bottomrule 
\end{tabular}

}}{
 \caption{Results on Charades-STA~\cite{charades-sta} for Temporal Video Grounding (TVG) task. \textit{LLaVA-ST} achieves the state-of-the-art performance.}
 \label{tab:tvg}
}
\capbtabbox{
 \resizebox{0.95\linewidth}{!}{
\setlength{\tabcolsep}{5 pt}
\renewcommand{\arraystretch}{1.05}

\setlength{\tabcolsep}{2.2pt}
\begin{tabular}{@{}l|ccc|ccc|cc@{}}
\toprule

\multirow{2}{*}{ Model} & \multicolumn{3}{c|}{RefCOCO} & \multicolumn{3}{c|}{RefCOCO+} & \multicolumn{2}{c}{RefCOCOg} \\
 & val & test-A & test-B & val & test-A & test-B & val-u & test-u \\

\midrule

\multicolumn{9}{c}{\it Generalist models w. full resolution input} \\
\midrule
\multicolumn{1}{l|}{\color{gray}Ferret v2-7B~\cite{you2023ferret}*}  & \color{gray} 92.8 & \color{gray} 94.7 & \color{gray} 88.7 & \color{gray} 87.4 & \color{gray}  {92.8} & \color{gray} 79.3 & \color{gray} 89.4 & \color{gray} 89.3  \\
\midrule
\multicolumn{9}{c}{\it Generalist models w/o full resolution input} \\
\midrule
\multicolumn{1}{l|}{OFA-L\cite{wang2022ofa}}                   & 80.0 & 83.7 & 76.4 & 68.3 & 76.0 & 61.8 & 67.6 & 67.6 \\
\multicolumn{1}{l|}{Shikra-7B\cite{chen2023shikra}} & 87.0 & 90.6 & 80.2 & 81.6 & 87.4 & 72.1 & 82.3 & 82.2 \\
\multicolumn{1}{l|}{GroundingGPT-7B\cite{groundinggpt}} & 88.0 & 91.6 & 82.5 & 81.6 & 87.2 & 73.2 & 81.7 & 82.0  \\
\multicolumn{1}{l|}{MiniGPT-v2-7B\cite{chen2023minigptv2}}      & 88.7 & 91.7 & 85.3 & 80.0 & 85.1 & 74.5 & 84.4 & 84.7 \\
\multicolumn{1}{l|}{Ferret-7B\cite{you2023ferret}}      & 87.5 & 91.4 & 82.5 & 80.8 & 87.4 & 73.1 & 83.9 & 84.8 \\
\multicolumn{1}{l|}{VistaLLM-7B\cite{pramanick2024jack}} & 88.1 & 91.5 & 83.0 & 82.9 & 89.8 & 74.8 & 83.6 & 84.4  \\
\multicolumn{1}{l|}{Elysium-7B\cite{elysium}} & 89.1 & 92.1 & 85.0 & 82.9 & 88.9 & 75.6 & 82.9 & 83.6  \\
\multicolumn{1}{l|}{Groma-7B\cite{ma2025groma}}                & 89.5 & 92.1 & \textbf{86.3} & 83.9	& 88.9 & 78.1 & 86.3 & 87.0  \\
\midrule
\multicolumn{1}{l|}{\textit{LLaVA-ST}-7B} & \textbf{90.1} & \textbf{93.2} & 85.0 & \textbf{86.0} & \textbf{91.3} & \textbf{78.8} & \textbf{86.7} & \textbf{87.4} \\
\bottomrule
\end{tabular}
}}{
 \caption{Performance on RefCOCO~\cite{refcoco}, RefCOCO+~\cite{refcoco} and RefCOCOg~\cite{refcocog} for Referring Expression Comprehension (REC). We report the accuracy with IoU Threshold set to 0.5.}
 \label{tab:rec}
}
\vspace{-1mm}
\end{floatrow}

\end{table*}

\begin{table*}[t]
  \centering
    \resizebox{\linewidth}{!}{
    \setlength{\tabcolsep}{6mm}{
        \begin{tabular}{l|cc|cc|ccccc|c}
    \toprule
    \multirow{2}{*}{Model} &\multicolumn{2}{|c}{MSVD-QA} &\multicolumn{2}{|c}{MSRVTT-QA}  &\multicolumn{6}{|c}{VCG-Bench}\\
    \cmidrule(lr){2-3} \cmidrule(lr){4-5} \cmidrule(lr){6-11}
    &Acc. &Score &Acc. &Score  &CI &DO &CU &TU &CO &Avg.\\
    \midrule 

    Video-LLaMA \citep{videollama}     & 51.6 & 2.5 & 29.6 & 1.8 & 1.96 & 2.18 & 2.16 & 1.82 & 1.79 & \cellcolor{gray!10}{1.98} \\  
    Video-ChatGPT \citep{video-chatgpt} & 64.9 & 3.3 & 49.3 & 2.8 & 2.50 & 2.57 & 2.69 & 2.16 & 2.20 & \cellcolor{gray!10}{2.42} \\
    GroundingGPT \cite{groundinggpt}          & 67.8    & 3.7   & 51.6    & 3.1    & - & - & - & - & - & -\\
    Momentor \citep{momentor}           & 68.9 & 3.6 & 55.6 & 3.0 & -    & -    & -    & -    & -    & -    \\
    MovieChat \citep{moviechat}         & 75.2 & 3.8 & 52.7 & 2.6 & 2.76 & 2.93 & 3.01 & 2.24 & 2.42 & \cellcolor{gray!10}{2.67} \\
    VTimeLLM \citep{vtimellm}           & -    & -   & -    & -   & 2.78 &  {3.10} & 3.40 & 2.49 & 2.47 & \cellcolor{gray!10}{2.85} \\

    LongVLM \citep{weng2025longvlm}             & 70.0 & 3.8 & 59.8 & 3.3 & 2.76 & 2.86 & 3.34 & 2.39 & 3.11 & \cellcolor{gray!10}{2.89}\\
    VideoChat2 \citep{mvbench}          & 70.0 &  {3.9} & 54.1 & 3.3 & 3.02 & 2.88 & 3.51 & 2.66 & 2.81 & \cellcolor{gray!10}{2.98} \\
    Chat-UniVi \citep{chatuniv}        & 65.0 & 3.6 & 54.6 & 3.1 & 2.89 & 2.91 & 3.46 & 2.89 & 2.81 & \cellcolor{gray!10}{2.99} \\
    LITA \citep{lita}                   & -    & -   & -    & -   & 2.94 & 2.98 & 3.43 & 2.68 &  {3.19} & \cellcolor{gray!10}{3.04} \\

    P-LLaVA-7B \citep{xu2024pllava}           & {76.6} & \textbf{4.1} & {62.0} & {3.5} & 3.21 & 2.86 & 3.62 & 2.33 & 2.93 & \cellcolor{gray!10}{3.12} \\
    ST-LLM \citep{st-llm}               & 74.6 &  {3.9} & {63.2} & 3.4 & 3.23 & 3.05 & {3.74} &  {2.93} & 2.81 & \cellcolor{gray!10}{3.15} \\
    VideoGPT+ \citep{videogpt+}          & -    & -   & -    & -    &  {3.27} & \textbf{3.18} & {3.74} & 2.83 & {3.39} & \cellcolor{gray!10}{{3.28}} \\
     Elysium \cite{elysium}          & 75.8    & 3.7   & \textbf{67.5}    & 3.2  & - & - & - & - & - & -\\

    Grounded-VideoLLM~\cite{groundedvideollm}       &\textbf{76.3} & {\textbf{4.1}} & 60.3 & \textbf{3.6} & \textbf{3.34} & 2.94 & 3.66 & \textbf{3.12} & 3.14 & \cellcolor{gray!10}{3.24} \\ 
    \midrule 

    \textit{LLaVA-ST}       & {75.9} & \textbf{4.1} & 59.0 & {3.5} & {3.29} & \textbf{3.18} & \textbf{3.80} &  {2.9} & \textbf{3.41} & \cellcolor{gray!10}{\textbf{3.32}} \\ 
    \bottomrule 
\end{tabular}

}
}
\caption{Results on Open-Ended VideoQA and VCG-Bench. VCG-Bench contains five aspects: Correctness of Information (CI), Detail Orientation (DO), Contextual Understanding (CU), Temporal Understanding (TU), and Consistency (CO).}
\label{tab:results_qa_vcg}
\end{table*}
\subsection{Training stategy}
To enhance the stability of the training process and improve the model's final performance, we partitioned the training data into three stages based on data quality and the granularity of visual-text alignment. 
As shown in Table~\ref{tab:data}, these stages are: content alignment, coordinate alignment, and multi-task instruction tuning.

\noindent{\textbf{\textit{S}1. Content alignment}}. We utilize video caption data~\cite{webvid,panda70m,intervid10m} in this stage to preliminarily achieve vision-and-language content alignment in the outputs of the STP module. In this process, we fix the LLM's parameters and only train the parameters within STP. Due to the absence of fine-grained multimodal understanding data, we set $\hat{\bm{\rho}}$ as $\bm{0}$ to minimize disturbances. STP's learning rate is $1\times 10^{-3}$.

\noindent{\textbf{\textit{S}2. Coordinate alignment}}. During the second stage of training, we utilize tasks requiring fine-grained visual understanding to enable the model to align spatiotemporal coordinate between visual and textual spaces. In this stage, the data primarily consists of fine-grained multimodal understanding datasets constructed via automatic generation methods. For tasks involving fine-grained spatial understanding, we introduce three types of tasks—REC, DGC, and RC—based on the GranD~\cite{glamm} dataset to train the alignment of spatial position embeddings, resulting in a total of $448k$ data samples. Simultaneously, for temporal fine-grained understanding in videos, we incorporate three types of tasks—TVG~\cite{vtimellm}, DVC~\cite{vtimellm,momentor,internvideo2}, and TR~\cite{vtimellm,intervid10m}—to align coordinate along the temporal dimension, with a total of $336$K data samples. In this stage, we apply a learning rate of $2 \times 10^{-4}$ to the STP and train the main network of LLM using LoRA~\cite{hu2022lora} with the same learning rate.

\noindent{\textbf{\textit{S}3. Multi-task instruction tuning}}. In the third phase, we introduce $39$ high-quality annotated datasets to endow the model with general capabilities of visual question answering and fine-grained multimodal understanding. We incorporate for spatial and temporal fine-grained understanding tasks and text-response tasks, like REC, TVG and Video QA as shown in Table~\ref{tab:data}. 
Additionally, we incorporate data involving spatial-temporal interleaved fine-grained understanding, \ie STVG, ELC, and SVG. We add new LoRA parameters to the stage-2 model, while employing the same training strategy as in stage-2.

\section{Experiments}

\subsection{Implements Details}
We use SigLIP-400M~\cite{zhai2023sigmoid} as $f_v$ with resolution of $384\times 384$. The pre-trained model is Llava-onevision 7B~\cite{llavaov}. The $M_w$, $M_h$, $M_t$ are $100$. The $k_1$ is $9$, $k_2$ is $3$ and $\sigma$ is 20. $N$ is $100$. $H_1$ is $27$ and $W_1$ is $27$. The training batch size is $384$ for stage-1 and 192 for stage-2 and stage-3. We train on ST-Align for 1 epoch. The $\alpha$ and $r$ of LoRA~\cite{hu2022lora} is 128 and 256. We train the \textit{LLaVA-ST} with $48$ A100 in $72$ hours. We use AdamW~\cite{loshchilov2017decoupled} as optimizer, with a cosine learning rate decay and a warm-up period.
\subsection{Main Comparisons}
\label{sec:exp:comparisions}

In the following sections, we conduct comprehensive experimental analysis of \textit{LLaVA-ST}. \textit{LLaVA-ST} achieves outstanding performance across 12 benchmarks for different multimodal understanding capability.  It is important to note that all our test results across different benchmarks are obtained using the same model by modifying the prompts.

\noindent{\textbf{Spatial-temporal interleaved Tasks}}. 
Unlike previous MLLMs, \textit{LLaVA-ST} can process spatiotemporal interleaved tasks end-to-end. Based on the benchmark of ST-Align (as detailed in \S\ref{sec:training}), we evaluate \textit{LLaVA-ST} on three tasks: STVG, ELC, and SVG.  We construct corresponding baselines based on GroundingGPT~\cite{groundinggpt}, employing a two-stage approach to accomplish these tasks.  Furthermore, we evaluate Grounded-VideoLLM~\cite{groundedvideollm} and VTimeLLM~\cite{vtimellm} on the ST-Align benchmark, focusing on metrics related to temporal localization and language generation.
\textit{LLaVA-ST} demonstrates significant improvements across these tasks compared to previous methods. In the STVG task, it achieves $43.8$ on m$_t$IoU outperforming Grounded-VideoLLM~\cite{groundedvideollm} by $10.8$ and achieves $22.8$ on m$_s$IoU, surpassing GroundingGPT~\cite{groundinggpt} by $13.6$. In the ELC task, \textit{LLaVA-ST} improves m$_t$IoU, m$_s$IoU, and METEOR over other model’s best performance by $27.3$, $19.7$ and $18.7$ respectively. Specifically, In the SVG task, it increases m$_s$IoU by $14.6$ compared to GroundingGPT~\cite{groundinggpt}. \textit{LLaVA-ST}'s end-to-end processing capability for spatial-temporal interleaved fine-grained understanding tasks offers substantial advantages in both inference speed and accuracy.

\noindent{\textbf{Temporal Video Grounding (TVG)}}. As shown in Table~\ref{tab:tvg}, 
We validate the TVG capabilities of \textit{LLaVA-ST} on the Charades-STA~\cite{charades-sta} benchmark. In the TVG task, the goal is to localize the start and end timestamps of events in videos based on language descriptions. \textit{LLaVA-ST} outperforms the previous MLLMs, Grounded-VideoLLM~\cite{groundedvideollm}, achieving an improvement of $6.9$ in mIoU. This result demonstrates MLLM's strong temporal localization capability.

\noindent{\textbf{Referring Expression Comprehension (REC)}}. We evaluate the fine-grained spatial understanding capabilities of \textit{LLaVA-ST} on standard REC datasets, \ie, RefCOCO~\cite{refcoco}, RefCOCO+~\cite{refcoco}, and RefCOCOg~\cite{refcocog}. Compared to the previous state-of-the-art model without full-resolution image inputs, Groma-7B~\cite{ma2025groma} , \textit{LLaVA-ST} achieves superior accuracy with the same model size. Notably, on the RefCOCO+~\cite{refcoco} dataset, \textit{LLaVA-ST} improves by $2.1$\% on the validation set and $2.4$\% on the test-A set. This demonstrates that \textit{LLaVA-ST} possesses strong fine-grained spatial understanding abilities. Furthermore, it demonstrates that the LLM can simultaneously process visual features of two different structures, $\hat{\bm{v}}$ and ${\hat{\bm{F}}_{s/t}}$.

\noindent{\textbf{Open-Ended Video QA}}. 
\textit{LLaVA-ST} achieved promising results on the Open-Ended Video QA task, which requires models to generate open-ended responses based on video content. Specifically, it attains the highest average score of $3.32$ on the VCG-Bench~\cite{video-chatgpt} dataset and demonstrates strong performance on the MSVD-QA~\cite{xu2017video} and MSRVTT-QA~\cite{xu2017video} benchmarks. These findings indicate that \textit{LLaVA-ST}'s support for various fine-grained multimodal understanding tasks does not compromise its performance on general multimodal question answering.

\begin{table*}
  \centering
    \resizebox{\linewidth}{!}{
    \setlength{\tabcolsep}{1.5mm}

 \begin{tabular}{l|c|cccccccccccccccccccc}
    \toprule
    Model &Avg. &AS &AP &AA &FA &UA &OE &OI &OS &MD &AL &ST &AC &MC &MA &SC &FP &CO &EN &ER &CI \\
    \midrule 
    VideoChatGPT \citep{video-chatgpt} & \cellcolor{gray!10}{32.7} & 23.5 & 26.0 & 62.0 & 22.5 & 26.5 & 54.0 & 28.0 & 40.0 & 23.0 & 20.0 & 31.0 & 30.5 & 25.5 & 39.5 & 48.5 & 29.0 & 33.0 & 29.5 & 26.0 & 35.5 \\
    VideoLLaMA \citep{videollama}     & \cellcolor{gray!10}{34.1} & 27.5 & 25.5 & 51.0 & 29.0 & 39.0 & 48.0 & 40.5 & 38.0 & 22.5 & 22.5 & 43.0 & 34.0 & 22.5 & 32.5 & 45.5 & 32.5 & 40.0 & 30.0 & 21.0 & 37.0 \\
    VideoChat \citep{videochat}        & \cellcolor{gray!10}{35.5} & 33.5 & 26.5 & 56.0 & 33.5 & 40.5 & 53.0 & 40.5 & 30.0 & 25.5 & 27.0 & 48.5 & 35.0 & 20.5 & 42.5 & 46.0 & 26.5 & 41.0 & 23.5 & 23.5 & 36.0 \\
    TimeChat \citep{timechat}          & \cellcolor{gray!10}{38.5} & 40.5 & 36.0 & 61.0 & 32.5 & 53.0 & 53.5 & 41.5 & 29.0 & 19.5 & 26.5 & 66.5 & 34.0 & 20.0 & 43.5 & 42.0 & 36.5 & 36.0 & 29.0 & 35.0 & 35.0 \\
    Video-LLaVA \citep{videollama}    & \cellcolor{gray!10}{43.0} & 46.0 & 42.5 & 56.5 & 39.0 & 53.5 & 53.0 & 48.0 &  {41.0} & 29.0 & 31.5 & 82.5 &  {45.0} & 26.0 & 53.0 & 41.5 & 33.5 & 41.5 & 27.5 & 38.5 & 31.5 \\
    P-LLaVA-7B \citep{xu2024pllava}          & \cellcolor{gray!10}{46.6} & 58.0 & 49.0 & 55.5 & 41.0 & 61.0 & 56.0 & 61.0 & 36.0 & 23.5 & 26.0 & 82.0 & 39.5 & 42.0 & 52.0 & 45.0 & 42.0 &  {53.5} & 30.5 &  {48.0} & 31.0 \\
    VideoChat2 \citep{mvbench}         & \cellcolor{gray!10}{51.1} & 66.0 & 47.5 & 83.5 & 49.5 & 60.0 & 58.0 & 71.5 &  \textbf{42.5} & 23.0 & 23.0 &  {88.5} & 39.0 & 42.0 & 58.5 & 44.0 &  {49.0} & 36.5 & 35.0 & 40.5 &  \textbf{65.5} \\
    ShareGPT4Video \citep{sharegpt4video} & \cellcolor{gray!10}{51.2} & 49.5 & 39.5 & 79.5 & 40.0 & 54.5 &  {82.5} & 54.5 & 32.5 &  \textbf{50.5} &  {41.5} & 84.5 & 35.5 &  {62.5} & 75.0 &  {51.0} & 25.5 & 46.5 & 28.5 & 39.0 & 51.5 \\
    ST-LLM \citep{st-llm}              & \cellcolor{gray!10}{54.9} & 66.0 & 53.5 &  {84.0} & 44.0 & 58.5 & 80.5 & 73.5 & 38.5 & 42.5 & 31.0 & 86.5 & 36.5 & 56.5 &  {78.5} & 43.0 & 44.5 & 46.5 &  {34.5} & 41.5 & 58.5 \\
    VideoGPT+ \citep{videogpt+}        & \cellcolor{gray!10}{ {58.7}} &  {69.0} &  {60.0} &  {83.0} &  {48.5} &  {66.5} &  {85.5} &  {75.5} & 36.0 &  {44.0} & 34.0 &  {89.5} & 39.5 &  {71.0} &  {90.5} & 45.0 &  {53.0} &  {50.0} & 29.5 &  {44.0} &  {60.0} \\
    {Grounded-VideoLLM}~\cite{groundedvideollm} &\cellcolor{gray!10}{ {59.4}} & {76.0} & \textbf{75.5} &77.0 & {48.0} & {67.5} & {85.5} & {77.0} &34.5 &39.5 & \textbf{59.5} &86.5 & \textbf{44.5} &60.5 &79.0 & \textbf{51.5} & {49.0} &46.0 & {35.0} &42.5 &54.0 \\
    \midrule 
    \textit{LLaVA-ST} &\cellcolor{gray!10}{ \textbf{64.2}} & \textbf{77.0} & {69.0} & \textbf{91.5} & \textbf{50.0} & \textbf{68.5} & \textbf{93.5} & \textbf{84.5} &40.0 &44.5 & {49.5} &\textbf{93.0} & {44.0} &\textbf{77.5} &\textbf{97.5} & {41.0} & \textbf{57.0} &\textbf{56.5} & \textbf{37.0} &\textbf{49.0} &63.0 \\
    \bottomrule 
\end{tabular}
}
\caption{Results on MVBench for multi-choice question answering.}
\label{tab:results_mvbench}
\end{table*}

\begin{table*}
    
\begin{floatrow}
\capbtabbox{
\resizebox{0.95\linewidth}{!}{
\setlength{\tabcolsep}{7 pt}
\begin{tabular}{l|c|cc|c}
\toprule
\multirow{2}{*}{Method} & RefCOCO (val)     & \multicolumn{2}{c|}{ST-Align (STVG)} & Charades-STA  \\
                        & Acc@0.5 & m$_t$IoU & m$_s$IoU            & mIoU          \\
\midrule
LLaVA-ST                &  \textbf{89.6}        & \textbf{43.5}  &  \textbf{21.0}            & \textbf{38.5}          \\
w/o LAPE                &     88.7         &   42.1    &    17.9              & 36.1           \\
w/o STP                 &   89.2       &  42.0     &    18.2              &  35.5           \\
\bottomrule
\end{tabular}
}}{
 \caption{Ablation studies about modules in \textit{LLaVA-ST}. The result on REC, STVG and TVG is evaluated on RefCOCO~\cite{refcoco}, ST-Align and Charades-STA~\cite{charades-sta} respectively. The proposed LAPE and STP modules can significantly enhance the performance of \textit{LLaVA-ST}.}
 \label{tab:tb1}
}
\capbtabbox{
 \resizebox{0.95\linewidth}{!}{
\setlength{\tabcolsep}{6 pt}
\begin{tabular}{ccc|ccc|ccc}
\toprule
\multicolumn{3}{c|}{Training Data} & \multicolumn{3}{c|}{RefCOCO}      & \multicolumn{3}{c}{ Charades-STA}  \\
T & S & T+S                       & val & test-A & test-B & R@0.3 & R@0.5            & mIoU          \\
\midrule
 $\checkmark$ &   &                           &        -    & -  &  -    & 48.7  &   32.0              & 34.6           \\
 $\checkmark$ & $\checkmark$  &               &     \textbf{89.8}  &\textbf{ 92.5}  & \textbf{85.1}   & 52.9  &   34.1              & 32.3           \\
 $\checkmark$ & $\checkmark$  & $\checkmark$  &     89.6  & 92.1 & \textbf{85.1}      & \textbf{58.0}  &  \textbf{39.8}    & \textbf{38.5}           \\
\bottomrule
\end{tabular}
}}{
 \caption{Ablation studies about training data. The result of REC and TVG is evaluated on RefCOCO~\cite{refcoco} and Charades-STA~\cite{charades-sta} respectively. T represents Temporal related data, S represent spatial related data and T+S represents data includes spatial temporal interleaved localization. We report the accuracy with IoU threshold set to 0.5 for RefCOCO.}
 \label{tab:tb2}
}
\end{floatrow}

\end{table*}

\noindent\textbf{Multi-Choice Video QA}. We also validate the performance of \textit{LLaVA-ST} on multi-choice video QA, using MVBench~\cite{mvbench}. Compared to previous methods, \textit{LLaVA-ST} achieves the current best average score, improving by $4.8$ average score over previous models. \textit{LLaVA-ST} demonstrated significant improvements compared to prior approaches. Furthermore, on tasks that demand high spatial-temporal detailed information caption,such as AA (Action Antonym), OE (Object Existence), and MA (Moving Attribute), \textit{LLaVA-ST} achieves significant improvements over previous MLLMs.

\subsection{Ablation Study}
In this section, we conduct ablation studies on the design of \textit{LLaVA-ST} to investigate the contributions of the LAPE and STP modules in Table~\ref{tab:tb1}. We also assess the impact of each stage in progressive training on model performance in Table~\ref{tab:tb2}. In the ablation experiments, we employ a subset of the data from ST-Align for training. To comprehensively evaluate the spatiotemporal fine-grained understanding abilities of the ablated model in all aspects, we separately validate its capabilities on REC, STVG, and TVG tasks using the three benchmarks: RefCOCO~\cite{refcoco}, ST-Align, and Charades-STA~\cite{charades-sta}.

\noindent\textbf{About STP}.  Compared to the complete model presented in the first row of Table~\ref{tab:tb1}, Excluding STP from the model and employing direct pooling to process video features results in a decrease of $1.5$\% in the model's mtIoU on the STVG task. Similarly, the mIoU on TVG decreases by $3$\%. This indicates that STP can significantly enhance the effectiveness of video features and is of critical importance for spatiotemporal fine-grained understanding tasks in videos.

\noindent\textbf{About LAPE}. In the third row of Table~\ref{tab:tb1}, we present the results of \textit{LLaVA-ST}  without LAPE. Compared to the full \textit{LLaVA-ST} model, the  model without LAPE exhibits a decrease of $0.9$ in Acc@$0.5$ on the RefCOCO~\cite{refcoco} val set. For the temporal fine-grained understanding task TVG, the model's mIoU decreases by $2.4$. Additionally, in the STVG task, the model's m$_t$IoU and m$_s$IoU decrease by $1.4$ and $3.1$, respectively. Notably, the accuracy degradation is more pronounced for video tasks, indicating that LAPE is critical for multimodal coordinate alignment when  the coordinate space is extended.

\noindent\textbf{About Training Data}.
In Table~\ref{tab:tb2}, we examine the impact of different training data on the performance of \textit{LLaVA-ST}. Adding spatial-related data to the model trained solely on temporal fine-grained understanding (comparing row 2 to row 1) results in a decrease of $2.3$ in TVG mIoU on the Charades-STA dataset. However, incorporating spatial-temporal interleaved data (row 3) leads to a significant increase of $3.9$ in mIoU compared to using only temporal-related data. Meanwhile, the accuracy on the RefCOCO validation set are on the same level. These results demonstrate that spatial-temporal interleaved data is crucial for the temporal localization capabilities of \textit{LLaVA-ST}. Spatial localization information from spatiotemporal interleaved data enables the model to learn key objects involved in video events, thereby enhancing temporal localization. Therefore, unifying spatiotemporal fine-grained understanding within a single modeling framework can improve the model's performance in temporal localization.

\section{Conclusion}

In this paper, we propose \textit{LLaVA-ST}, a model capable of simultaneously processing temporal and spatial dimensions for fine-grained multi-modal understanding tasks.  
In terms of model architecture, we introduce LAPE, which explicitly embedding coordinate special tokens in the text space into the visual features as positional embeddings. Furthermore, we propose STP, which employs a point-to-region attention mechanism to compress temporal and spatial resolutions separately.
To aid future MLLMs in improving their capabilities in fine-grained multi-modal understanding, we propose a dataset called ST-Align and proposed a progressive training strategy to learn content alignment, coordinate alignment and multi-task ablility sequencially. Experimental results demonstrate that \textit{LLaVA-ST} exhibits strong fine-grained understanding capabilities across multiple benchmarks.
Furthermore, ablation experiments confirm the effectiveness of each module in our model and demonstrate the advantages of incorporating fine-grained spatiotemporally interleaved data in the proposed ST-Align method.

{
    \small
    \bibliographystyle{ieeenat_fullname}
    \bibliography{main}
}

\clearpage
\setcounter{page}{1}
\maketitlesupplementary

\section{Spatial Temporal Packer}
\begin{figure}[h]
    \centering
    \includegraphics[width=0.7\linewidth]{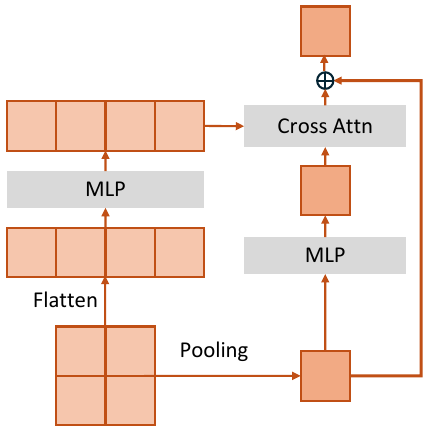}
    \caption{The architecture of $\text{packer}_s$, and $\text{packer}_t$ shares the similar architecture.}
    \label{fig:stp}
\end{figure}

As shown in \cref{fig:stp}, it illustrates the calculation process of a patch feature in $\text{packer}_s$. After obtaining the region feature and the pooled feature, each is processed through an MLP, followed by cross-attention computation. The pooled feature is then added to the output feature through a residual connection. $\text{packer}_t$ share the similar architecture but with temporal pooling dimension.
\section{More Ablations}
\begin{table}[]
    \centering
    \resizebox{\linewidth}{!}{
\begin{tabular}{l|c|cc|c}
\toprule
\multirow{2}{*}{Training Data} & RefCOCO (val) & \multicolumn{2}{c|}{ST-Align (STVG)} & Charades-STA  \\
                                & ACC@0.5         & m$_t$IoU             & m$_s$IoU            & mIoU          \\
\hline
S.1 + S.3               &      88.4   &  37.5     &    16.6              &      33.3         \\
 S.1 + (S.2 + S.3)          & \textbf{88.7}  &     41.2    &    18.3          &    \textbf{35.0}           \\
 S.1 + S.2 + S.3            & 88.6 &    \textbf{41.5}   &    \textbf{21.2}           &     {33.3}          \\

\bottomrule
\end{tabular}}
    \caption{Ablation study about training stage. (S.2 + S.3) indicates mix training data of S.2 and S.3.}
    %Ablation study about training stage. (S.2 + S.3) indecates mix training data of S.2 and S.3.
    \label{tab:supdata}
\end{table}
\noindent\textbf{About training stage.}

We investigated the impact of different training stage partitions on the performance of \textit{LLaVA-ST}. We experimented with omitting S.2 and with a mixed training strategy combining S.2 and S.3 data on the model pretrained on stage 1. We use a 25\% subset of all data for training.
As detailed in \cref{tab:supdata}, indicate that excluding Stage 2 leads to a significant decrease in model performance, particularly affecting the spatial fine-grained understanding in videos. Specifically, the m$_s$IoU for STVG dropped by 4.6. Moreover, mixing data from Stage 2 and Stage 3 also resulted in a performance decline, with a decrease of 2.9 in m$_s$IoU. However, the accuracy changes for Temporal Video Grounding (TVG) on the charades-STA and ST-Align datasets were minimal.
These findings suggest that segmenting Stage 2 can reduce the difficulty direct spatial-temporal localization, thereby enhancing the effectiveness of the progressive training strategy.

\begin{table}[]
    \centering
    \resizebox{\linewidth}{!}{
\begin{tabular}{cc|c|cc|c}
\toprule
\multicolumn{2}{c|}{Model}     & \multirow{3}{*}{token\#}  & \multicolumn{2}{c|}{ST-Align (STVG)} & Charades-STA  \\
$\text{packer}_t$ & $\text{packer}_s$ &                           & m$_t$IoU & m$_s$IoU            & mIoU          \\
\hline
25$\times$9$\times$9               & 0       & 2025   & 36.2      &      8.3            &    8.8           \\
0       &     100$\times$5$\times$5           & 2500   &  42.5    &       18.9            &    36.7      \\     20$\times$9$\times$9          & 100$\times$3$\times$3  & 2520  & \textbf{43.5}      &      \textbf{21.0}            &    \textbf{38.5}            \\ 
\bottomrule
\end{tabular}}
    \caption{The ablation study about $\text{packer}_s$ and $\text{packer}_t$. We display the feature resolutions of the outputs of packer$_s$ and packer$_t$ in spatial and temporal dimensions.}
    \label{tab:supstp}
\end{table}
\noindent\textbf{About STP}.
We validate the effectiveness of compressing the temporal and spatial dimensions separately in STP.
To this end, we remove either the packer$_s$ or packer$_t$ module from STP while adjusting the feature resolution to maintain a token count similar to the original STP, as shown in \cref{tab:supstp}.  we uses a 50\% subset of all fine-grained understanding data for training.
 When retaining only the spatially fine-grained features from packer$_t$, the TVG accuracy on Charades-STA drops significantly, from 38.5 to 8.8, highlighting the critical role of higher temporal resolution in temporal fine-grained understanding.
Meanwhile, the spatial localization of individual video frames also requires to distinguish the differences between objects in each frame, which is also significantly affected by the lower resolution in temporal dimension, leading to a marked decrease in m$_s$IoU performance on the STVG task, dropping sharply from 21.0 to 8.3.
Conversely, removing the spatially fine-grained features from packer$_t$ results in a notable decline in accuracy for both STVG and TVG. 
The results indicate that both the spatially fine-grained features from packer$_t$ and temporally fine-grained features from packer$_s$ are significant for fine-grained spatiotemporal understanding tasks. Retaining only one branch leads to a decline in performance.

\section{ST-Align Data Construction}
In GranD, we sample 350K images that have Dense Grounded Caption (DGC) annotations, \ie, providing descriptions for each object within the images. Based on the correspondence between language description and bounding box phrases, we extended the its annotations to enrich the REC and REG datasets. \cref{fig:supgrand} illustrates the structure of the instruction prompts for DGC, along with the generated REC and REG. For the segmentation data in Muse, we utilize the bounding rectangles of segmented regions as object bounding boxes, also filtering out some low-quality descriptions. Additionally, for spatial-temporal interleaved tasks, we constructed three task types—STVG, ELC, and SVG—by leveraging the correspondence between the spatiotemporal tubes and their linguistic descriptions from VidSTG~\cite{Vidstg}. The detailed construction of these instruction prompts is shown in \cref{fig:supstvg}.

\begin{figure*}[t] 
\centering

\begin{tcolorbox}[colback=gray!10, colframe=black, text width=0.9\textwidth, title={Box 1: Prompt for instruction data 
generated from  GranD  in S. 2}]
\hypertarget{box4}{}
\textbf{Dense Image Caption:}

$Q_{D1}$: Could you please give me a detailed description of the image? Please respond with interleaved bounding boxes for the corresponding parts of the answer. \\
$Q_{D2}$: Can you provide a thorough description of this image? Please output with interleaved bounding boxes for the corresponding phrases. \\
$Q_{D3}$: Please describe in detail the contents of the image. Please respond with interleaved bounding boxes for the corresponding parts of the answer. \\
$Q_{D4}$: Could you give a comprehensive explanation of what can be found within this picture? Please output with interleaved bounding boxes for the corresponding phrases.\\
$Q_{D5}$: Could you give me an elaborate explanation of this picture? Please respond with interleaved bounding boxes for the corresponding phrases. \\
$Q_{D6}$: Could you provide me with a detailed analysis of this photo? Please output with interleaved bounding boxes for the corresponding parts of the answer.\\

% \tcblower

\textbf{Referring Expression Comprehension:}

$Q_{R1}$: In this image, where is \textcolor{blue}{$<object>$} located? \\
$Q_{R2}$: Can you identify the position of \textcolor{blue}{$<object>$} within this image? \\
$Q_{R3}$: Could you indicate the region where \textcolor{blue}{$<object>$} is located in this image? \\
$Q_{R4}$: Please describe the location of \textcolor{blue}{$<object>$} in this image. \\
$Q_{R5}$: Where can I find \textcolor{blue}{$<object>$} in this image? \\
$Q_{R6}$: What part of this image contains \textcolor{blue}{$<object>$}? \\
$Q_{R7}$: Where does \textcolor{blue}{$<object>$} appear in this image? \\
$Q_{R8}$: Where is \textcolor{blue}{$<object>$} situated within this image? \\

% \tcblower

\textbf{Region Caption:}

$Q_{C1}$: What happened about the subject/object within the specified region \textcolor{blue}{$<box>$}? \\
$Q_{C2}$: Can you identify the event about the subject/object within the region \textcolor{blue}{$<box>$}? \\
$Q_{C3}$: Describe the event about subject/object located within the region \textcolor{blue}{$<box>$}. \\
$Q_{C4}$: Can you describe the object within the region \textcolor{blue}{$<box>$}? \\
$Q_{C5}$: What can you deduce about the object in the region \textcolor{blue}{$<box>$}? \\
$Q_{C6}$: Identify the specific object within the region \textcolor{blue}{$<box>$}. \\
$Q_{C7}$: Describe the object located within the region \textcolor{blue}{$<box>$}. \\

\end{tcolorbox}
\label{fig:supgrand}
\caption{Prompt of instruction data generated from GranD.\textcolor{blue}{$<box>$} indicates the bounding box of the object and \textcolor{blue}{$<object>$} represents the corresponding language of the object.}
\end{figure*}

\begin{figure*}[t] % 使用figure*环境创建双栏的浮动体
\centering

\begin{tcolorbox}[colback=gray!10, colframe=black, text width=0.9\textwidth, title={Box 2: Prompt for Spatiotemporal Interleaved Tasks}]
\hypertarget{box4}{}
\textbf{Temporal Related Question}

$Q_{T1}$: When does  \textcolor{blue}{$<event>$} occur in the video? \\
$Q_{T2}$: At which time interval in the video can we see  \textcolor{blue}{$<event>$} occurring? \\
$Q_{T3}$: During which time is  \textcolor{blue}{$<event>$} happening in the video? \\
$Q_{T4}$: Tell me the timestamp when  \textcolor{blue}{$<event>$} happened in the video. \\
$Q_{T5}$: At what time does  \textcolor{blue}{$<event>$} take place in the video? \\
$Q_{T6}$: At what point in the video can we observe  \textcolor{blue}{$<event>$} taking place?\\

\textbf{Spatial Related Question}

$Q_{S1}$: Where is the corresponding subject/object located? \\
$Q_{S2}$: Can you identify the position of the corresponding subject/object within this video? \\
$Q_{S3}$: Please describe the location of the corresponding subject/object in this video. \\

\textbf{Task Instruction}

$Q_{I1}$: Please firstly give the timestamps, and then give the spatial bounding box corresponding to each timestamp in the time period. \\
$Q_{I2}$: Please firstly give the end timestamp, then give the event associated with the object/subject, finally give the spatial bounding box corresponding to each timestamp in the time period. \\
$Q_{I3}$: Please give the spatial bounding box corresponding to each timestamp in the time period. \\

\textbf{Temporal Span}

$Temp_{T1}$: During the span of \{\textcolor{blue}{$s_i$},\textcolor{blue}{$e_i$}\} \\
$Temp_{T2}$: In the time range \{\textcolor{blue}{$s_i$},\textcolor{blue}{$e_i$}\} \\
$Temp_{T3}$: During the period \{\textcolor{blue}{$s_i$},\textcolor{blue}{$e_i$}\} \\
$Temp_{T4}$: Within the time frame of \{\textcolor{blue}{$s_i$},\textcolor{blue}{$e_i$}\} \\
$Temp_{T5}$: In the time period \{\textcolor{blue}{$s_i$},\textcolor{blue}{$e_i$}\} \\
$Temp_{T6}$: During \{\textcolor{blue}{$s_i$},\textcolor{blue}{$e_i$}\} \\
$Temp_{T7}$: Between \{\textcolor{blue}{$s_i$},\textcolor{blue}{$e_i$}\} \\
$Temp_{T8}$: At \{\textcolor{blue}{$s_i$},\textcolor{blue}{$e_i$}\} \\

\textbf{Spatial-Temporal Tube}

$Temp_{S}$: \textcolor{blue}{$s_i$}: \textcolor{blue}{$<box_i>$}, \textcolor{blue}{$s_{i+1}$}: \textcolor{blue}{$<box_{i+1}>$}, ..., \textcolor{blue}{$s_{i+x}$}: \textcolor{blue}{$<box_{i+x}>$}\\
\tcblower

\textbf{Spatial Temporal Video Grounding}

\textcolor{red}{$Q$}: \textcolor{blue}{$Q_{Ti}$} + \textcolor{blue}{$Q_{Si}$} + \textcolor{blue}{$Q_{I1}$} \textcolor{red}{$A$}: \textcolor{blue}{$Temp_{Ti}$} + \textcolor{blue}{$Temp_{S}$} \\

\textbf{Event Localization and Captioning}

\textcolor{red}{$Q$}: Start at \textcolor{blue}{$s_i$} + \textcolor{blue}{$Q_{Ci}$} + \textcolor{blue}{$Q_{Si}$} + \textcolor{blue}{$Q_{I2}$}  \textcolor{red}{$A$}: End at \textcolor{blue}{$e_i$} + \textcolor{blue}{$<event>$} +\textcolor{blue}{$Temp_{S}$} \\

\textbf{Spatial Video Grounding}

\textcolor{red}{$Q$}: \textcolor{blue}{$Temp_{Ti}$} + \textcolor{blue}{$<event>$}  + \textcolor{blue}{$Q_{Si}$} + \textcolor{blue}{$Q_{I3}$}  \textcolor{red}{$A$}: \textcolor{blue}{$Temp_{S}$} \\

\end{tcolorbox}
\caption{Prompt of instruction data for spatial-temporal interleaved tasks. \textcolor{blue}{$<event>$} denotes the linguistic description of the event portrayed in the video. The instructions and responses for STVG, ELC, and SVG are derived from a synthesis of templates addressing temporally related question, spatially oriented question, instruction, temporal span, and spatial-temporal tube. }
\label{fig:supstvg}
\end{figure*}

\section{Evaluation Metrics}

In Spatial Temporal Interleaved tasks, the primary metrics used to evaluate model performance are tIoU and sIoU. The tIoU measures the IoU between the time intervals predicted by the model and the ground truth time spans. Meanwhile, the sIoU calculates the average value of the bounding boxes within the overlapping regions of the predicted and ground truth time intervals. tIoU@0.5 represents the proportion of tIoU values exceeding 50\%. Additionally, m$_t$IoU denotes the average value of tIoU. Similarly, sIoU@0.5 and m$_s$IoU are calculated using analogous principles.

\section{Implementation Details for Baselines}
In \ref{sec:exp:comparisions}, we present the performance of other MLLMs~\cite{groundedvideollm,groundinggpt,vtimellm} on three tasks within ST-Align. This section will detail the specific implementations of these models.
To enable GroundingGPT to achieve STVG capabilities, we first use event descriptions to locate the temporal span of the video. Then, within the identified time range, we perform frame-by-frame spatial localization of the target. Similarly, for the ELC task, we use the following prompt to locate the temporal range and simultaneously generate the event description:

\noindent\texttt{Start at frame $<$start\_frame\_idx$>$, what happened about the subject/object within the specified region$<$box$>$? When did this event occur in the video? Ouput the temporal grounding, and output the event caption. Output your answer in this form: \{'caption':'the description of the event','tvg':\{t1,t2\}\}}
Based on the determined temporal range, we subsequently perform frame-by-frame object localization. 
For the SVG task, the target object's localization results are generated frame by frame using the event description.

For Grounded-VideoLLM, VtimeLLM, the following prompt was selected after multiple experiments to generate the ELC results:

\noindent\texttt{"Start at time $<$start\_frame\_idx$>$, what happened about video? Ouput end frame, and output the event caption. Output your answer in this form: \{\{'caption':'the description of the event','end frame': x\}\}"}

\section {Visualized Results}
We visualized various capabilities of LLaVA-ST, including spatial-temporal fine-grained understanding tasks for videos (\cref{fig:supstvg}, \cref{fig:supelc}, \cref{fig:supsvg}), temporal fine-grained understanding tasks for videos (\cref{fig:supstvg}, \cref{fig:supvtr}), and fine-grained understanding tasks (\cref{fig:supvdic}, \cref{fig:supvrec}) for images. Additionally, LLaVA-ST also demonstrates the ability to comprehend general language instructions and engage in multi-turn dialogues (\cref{fig:supvqa1}, \cref{fig:supvqa2}).
\begin{figure*}[ht]
    \centering
    \includegraphics[width=0.9\linewidth]{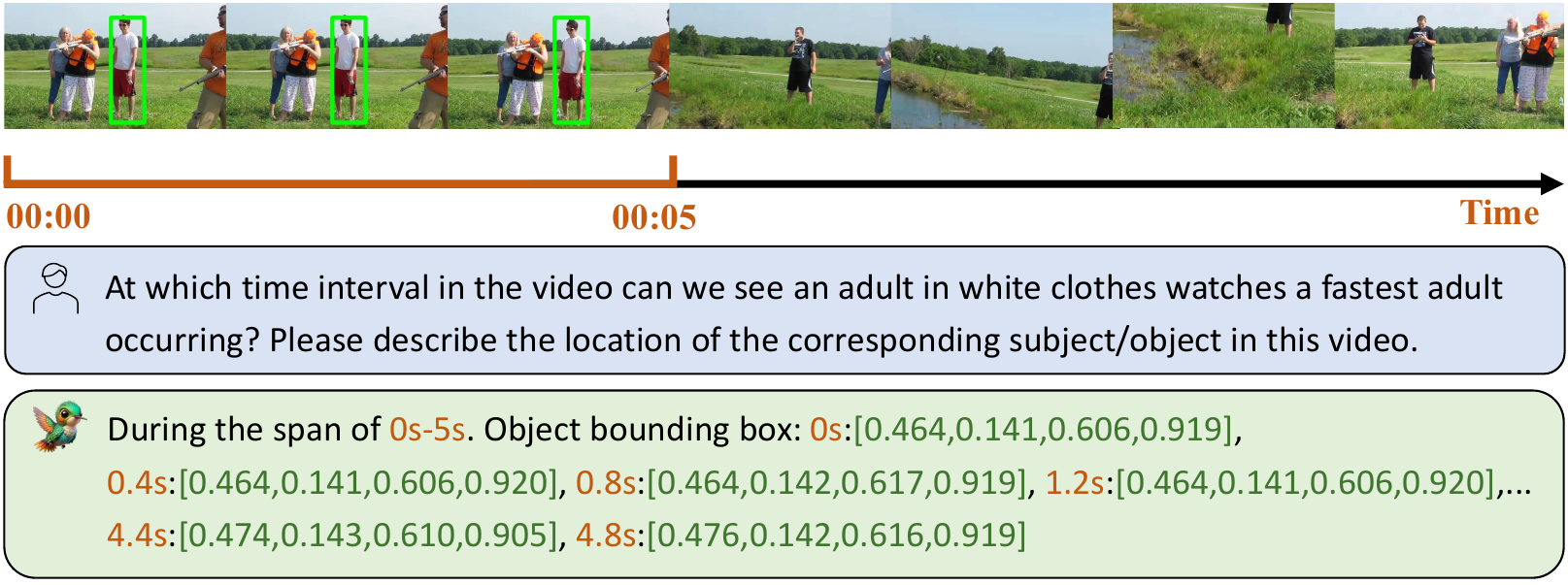}
    \caption{Visualized result of spatial-temporal video grounding.}
    \label{fig:supstvg}
\end{figure*}

\begin{figure*}[ht]
    \centering
    \includegraphics[width=0.9\linewidth]{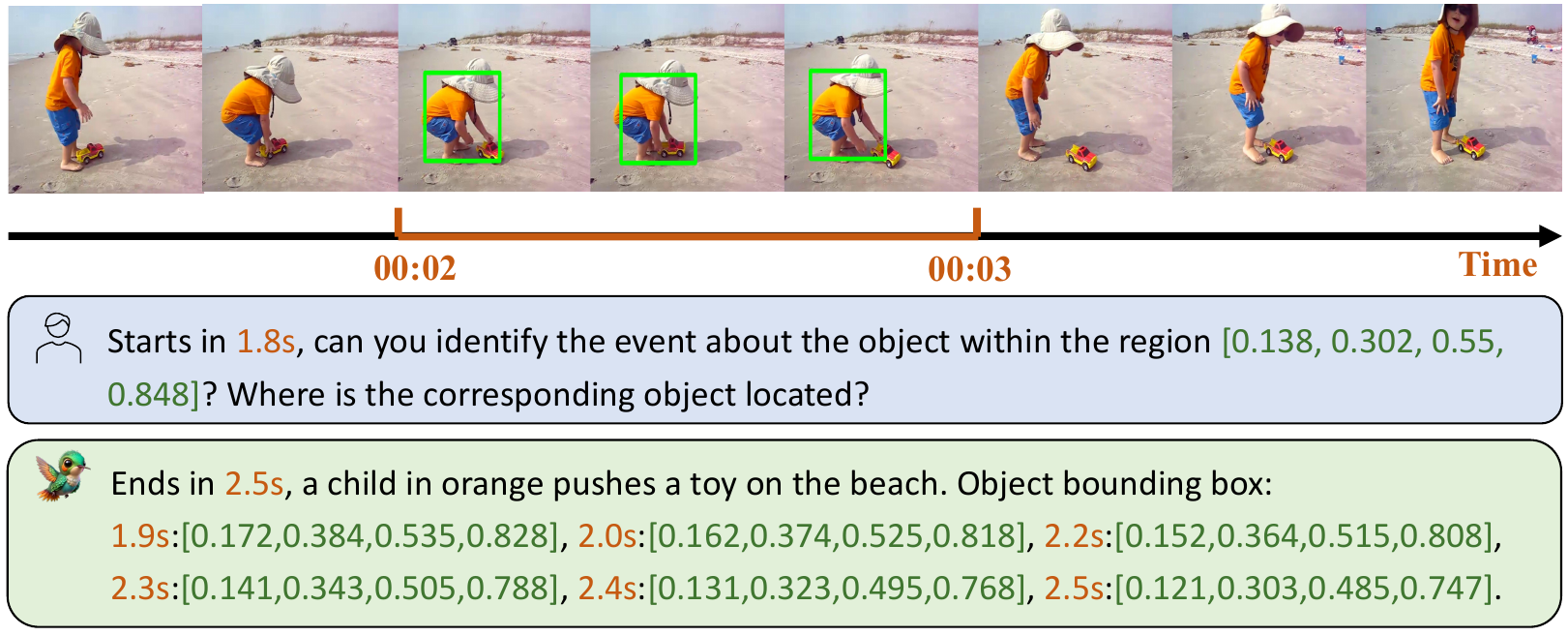}
    \caption{Visualized result of event localization and captioning.}
    \label{fig:supelc}
\end{figure*}

\begin{figure*}[ht]
    \centering
    \includegraphics[width=0.9\linewidth]{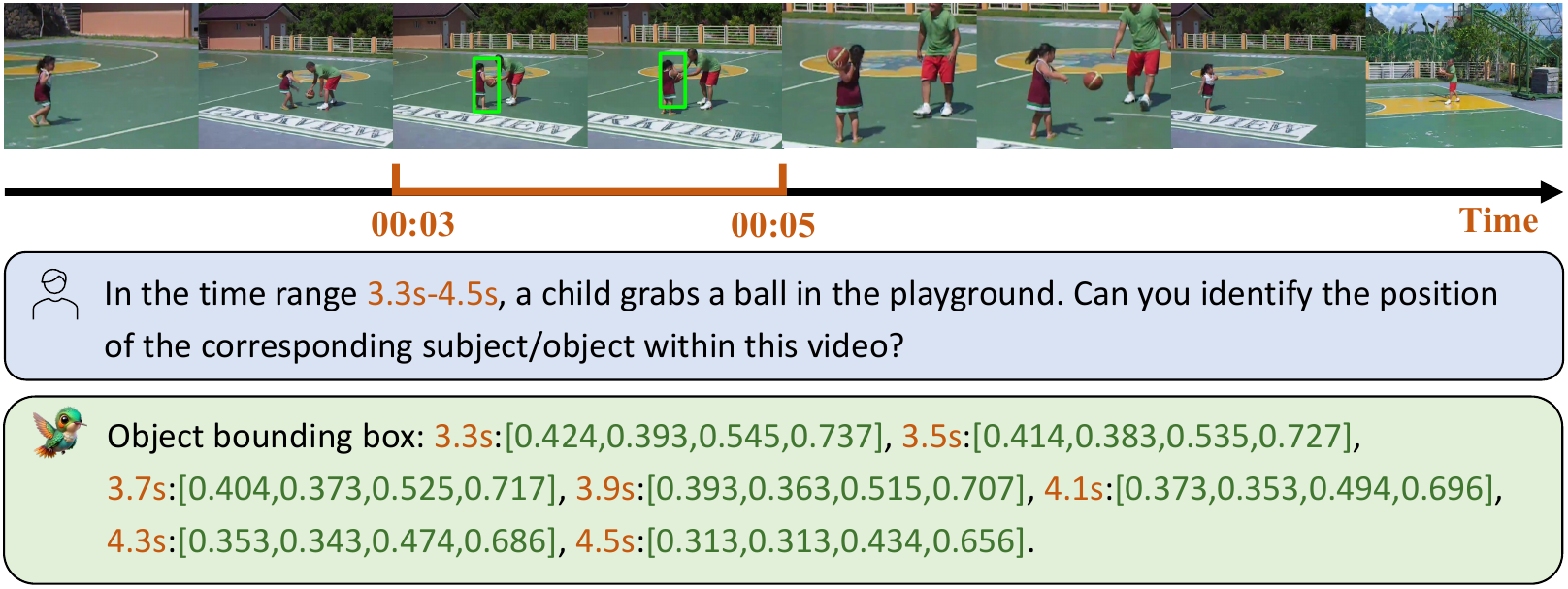}
    \caption{Visualized result of spatial video grounding.}
    \label{fig:supsvg}
\end{figure*}

\begin{figure*}[ht]
    \centering
    \includegraphics[width=0.9\linewidth]{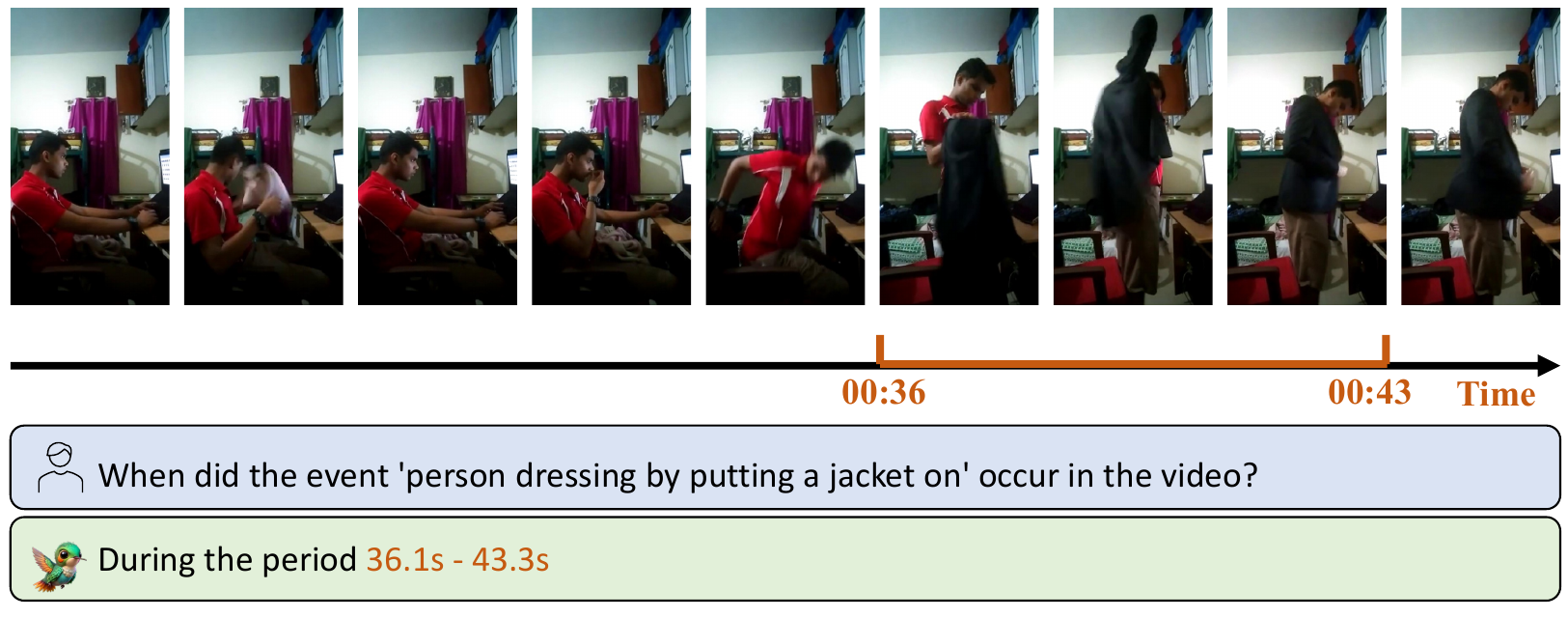}
    \caption{Visualized result of temporal video grounding.}
    \label{fig:supvtvg}
\end{figure*}

\begin{figure*}[ht]
    \centering
    \includegraphics[width=0.9\linewidth]{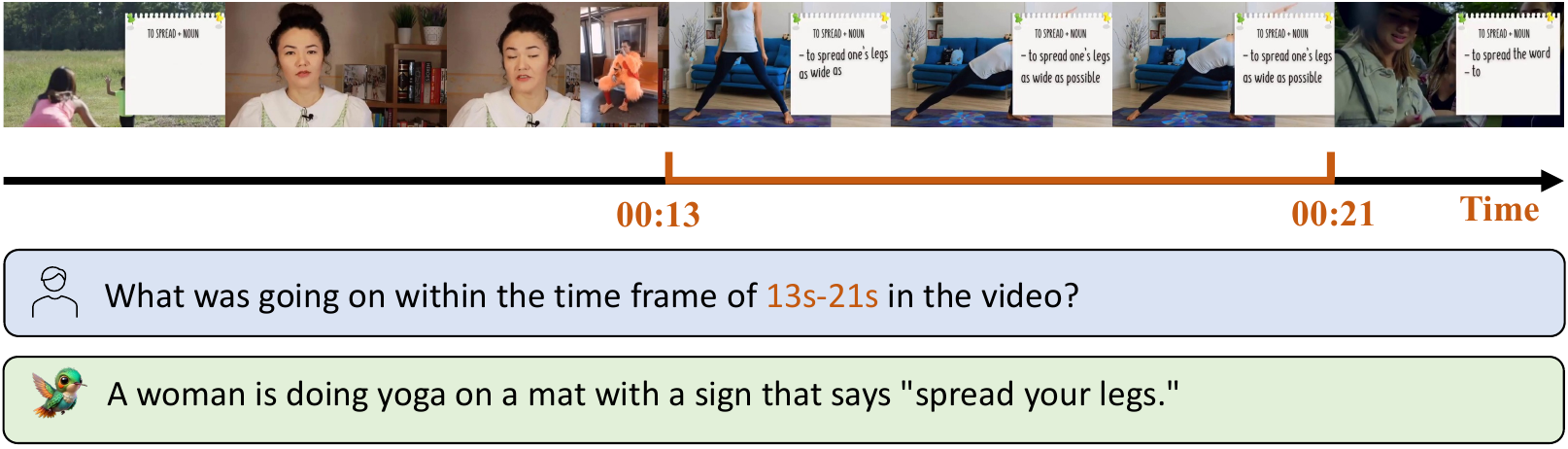}
    \caption{Visualized result of temporal referring.}
    \label{fig:supvtr}
\end{figure*}

\begin{figure*}[ht]
    \centering
    \includegraphics[width=0.75\linewidth]{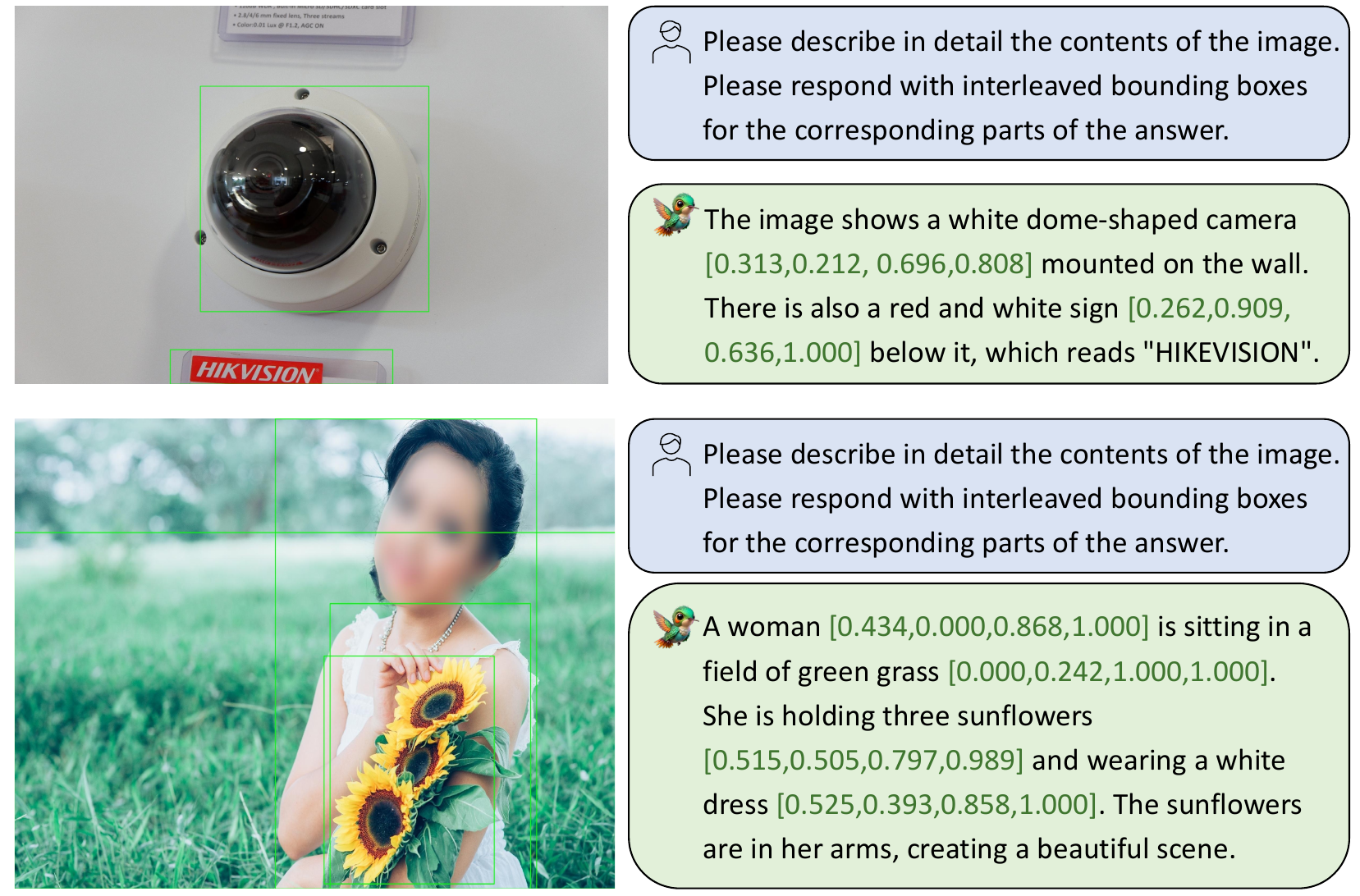}
    \caption{Visualized result of dense grounded captioning.}
    \label{fig:supvdic}
\end{figure*}
\begin{figure*}[h]
    \centering
    \includegraphics[width=0.9\linewidth]{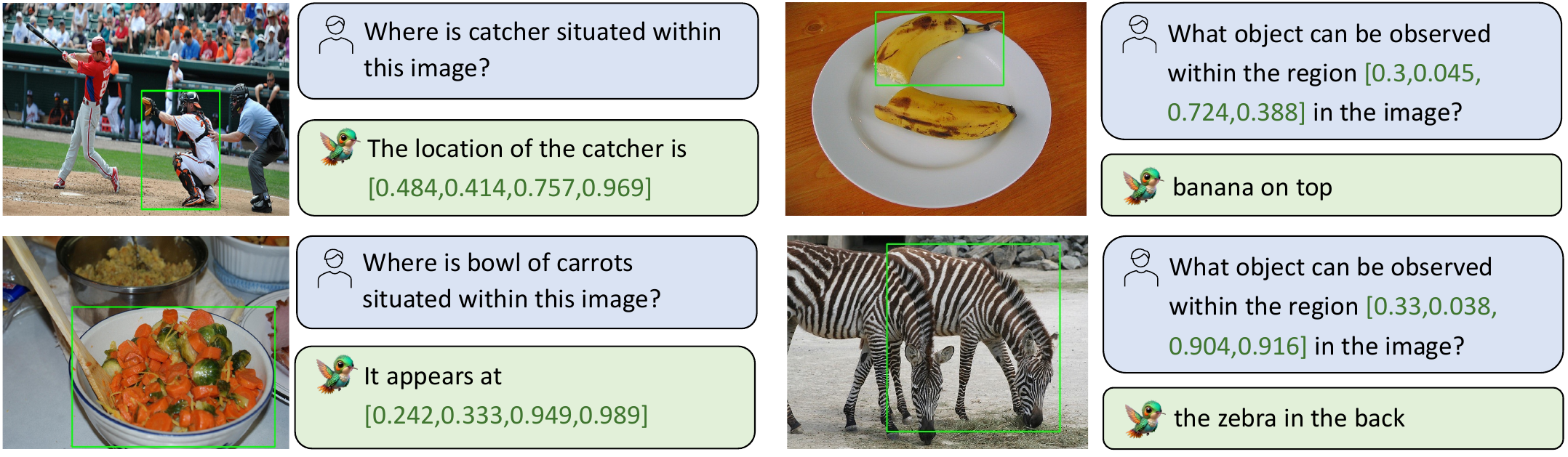}
    \caption{Referring expression comprehension and region caption visualize result.}
    \label{fig:supvrec}
\end{figure*}

\begin{figure*}[ht]
    \centering
    \includegraphics[width=0.9\linewidth]{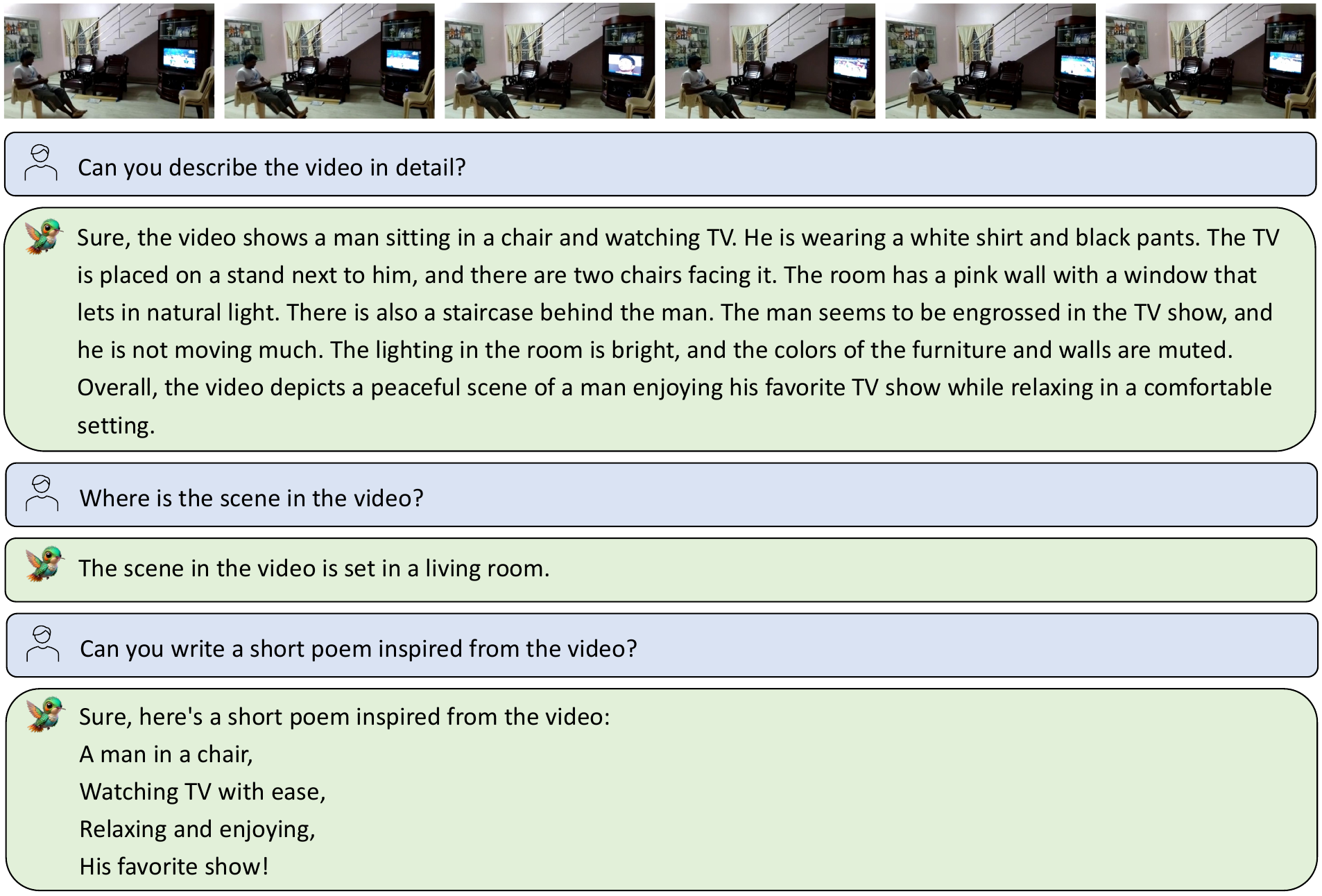}
    \caption{Example for multi-turn open-ended video QA of \texttt{LLaVA-ST}.}
    \label{fig:supvqa1}
\end{figure*}
\begin{figure*}[ht]
    \centering
    \includegraphics[width=0.9\linewidth]{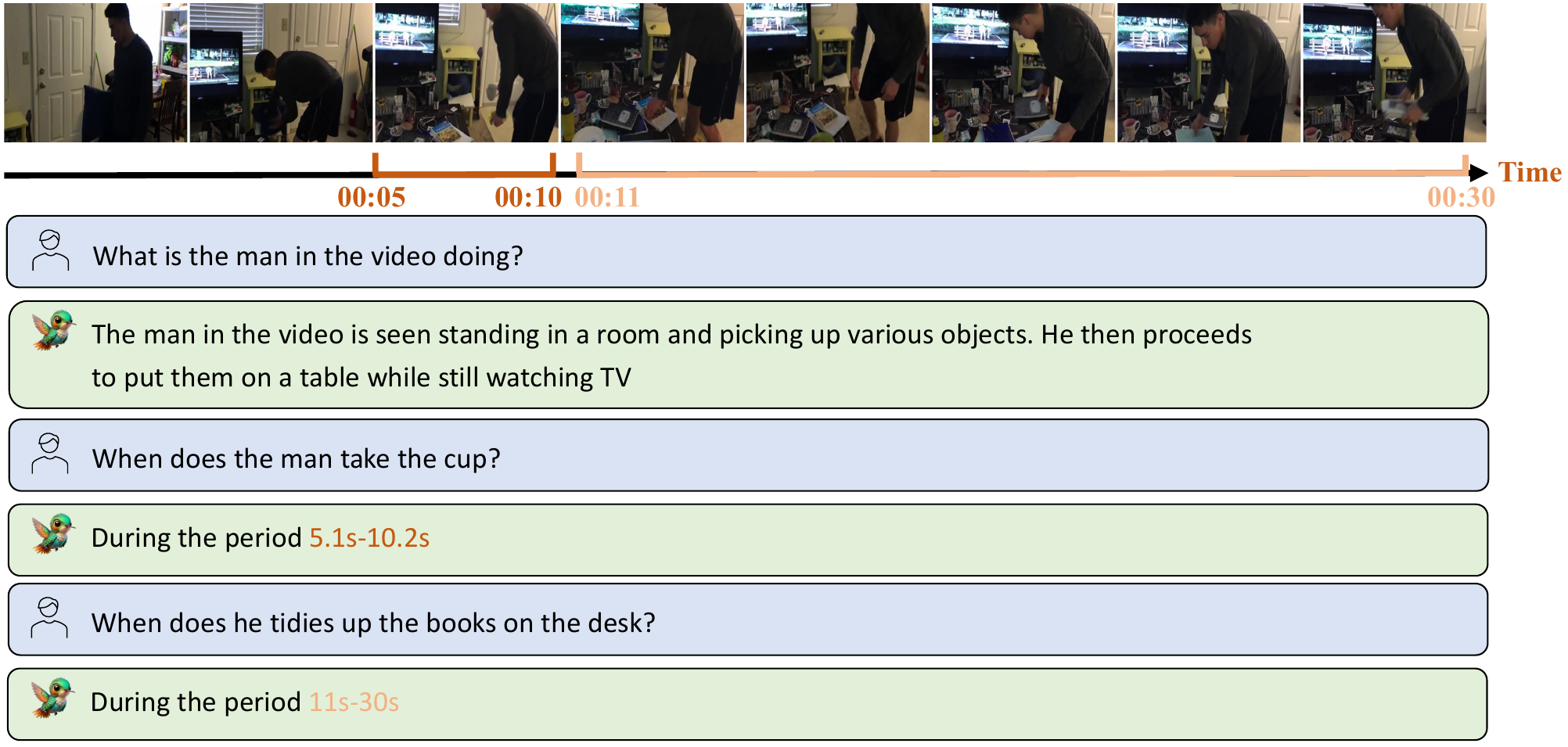}
    \caption{Example for multi-turn open-ended video QA of \texttt{LLaVA-ST}.}
    \label{fig:supvqa2}
\end{figure*}

\end{document}